%% file: han_567.tex
\documentclass[accepted]{uai2023} % after acceptance, for a revised
\usepackage[american]{babel}
\usepackage{natbib} % has a nice set of citation styles and commands
\usepackage{mathtools} % amsmath with fixes and additions
\usepackage{booktabs} % commands to create good-looking tables
\usepackage{tikz} % nice language for creating drawings and diagrams

\input{math_comands.tex}

\usepackage{comment}
\usepackage{algorithm}
\usepackage{algpseudocode}
\usepackage{subcaption}
\usepackage{multirow}
\usepackage{pifont}

\usepackage{amsmath}
\usepackage{amssymb}
\usepackage{mathtools}
\usepackage{amsthm}

\theoremstyle{plain}
\newtheorem{theorem}{Theorem}[section]

\newtheorem{lemma}[theorem]{Lemma}
\newtheorem{corollary}[theorem]{Corollary}
\theoremstyle{definition}

\newtheorem{assumption}[theorem]{Assumption}
\newtheorem{remark}[theorem]{Remark}

\newcommand{\trr}{\textcolor{black}}

\title{On the Convergence of Continual Learning with Adaptive Methods}

\author[1]{\href{mailto:<seungyubhan@snu.ac.kr>?Subject=On the convergence of continual learning}{Seungyub Han}{}}
\author[1]{Yeongmo Kim}
\author[1]{Taehyun Cho}
\author[1]{Jungwoo Lee}
\affil[1]{%
    Seoul National University\\
    Seoul, Republic of Korea
}
\begin{document}
\maketitle

\begin{abstract}
    One of the objectives of continual learning is to prevent catastrophic forgetting in learning multiple tasks sequentially,
    and the existing solutions have been driven by the conceptualization of the plasticity-stability dilemma.
    However, the convergence of continual learning for each sequential task is less studied so far.
    In this paper, we provide a convergence analysis of memory-based continual learning with stochastic gradient descent
    and empirical evidence that training current tasks causes the cumulative degradation of previous tasks.
    We propose an adaptive method for nonconvex continual learning (NCCL), which adjusts step sizes of both previous and current tasks with the gradients.
    The proposed method can achieve the same convergence rate as the SGD method when the catastrophic forgetting term which we define in the paper is suppressed at each iteration.
    Further, we demonstrate that the proposed algorithm improves the performance of continual learning over existing methods for several image classification tasks.
\end{abstract}

%%%%%%%%%%%%%%%%%%%%%%%%%%%%%%%%%%%%%%%%%%%%%%%%%%%%%%%%%%%%%% 
%%                   main body                              %%
%%%%%%%%%%%%%%%%%%%%%%%%%%%%%%%%%%%%%%%%%%%%%%%%%%%%%%%%%%%%%%

\section{Introduction}
\label{sec:intro}

Learning new tasks without forgetting previously learned tasks is a key aspect of artificial intelligence to be as versatile as humans.
Unlike the conventional deep learning that observes tasks from an i.i.d. distribution, continual learning train sequentially a model on a non-stationary stream of data \citep{DBLP:phd/dnb/Ring95, DBLP:conf/iros/Thrun94a}.
The continual learning AI systems struggle with catastrophic forgetting when the data access of previously learned tasks is restricted \citep{DBLP:journals/neco/FrenchC02}.
%%%% although
Although novel continual learning methods successfully learn the non-stationary stream sequentially, studies on the theoretical convergence analysis of both previous tasks and a current task have not yet been addressed.
In this line of research, nonconvex stochastic optimization problems have been well studied on a single task to train deep neural networks and prove theoretical guarantees of good convergence.

%%%%%%%%%%%%
Previous continual learning algorithms have introduced novel methods such as a replay memory to store and replay the previously learned examples \citep{lopez2017gradient,aljundi2019gradient,DBLP:conf/iclr/ChaudhryRRE19},
regularization methods that penalize neural networks \citep{kirkpatrick2017overcoming, zenke2017continual},
Bayesian methods that utilize the uncertainty of parameters or data points \citep{DBLP:conf/iclr/NguyenLBT18,DBLP:conf/iclr/EbrahimiEDR20},
and other recent approaches \citep{DBLP:conf/iclr/YoonYLH18,lee2019overcoming}.
The study of continual learning in Bayesian frameworks formulate a trained model for previous tasks parameter into an approximate posterior to learn a probabilistic model which have empirically good performance on entire tasks.
However, Bayesian approaches can fail in practice and it can be hard to analyze the rigorous convergence due to the approximation. 
The memory-based methods are more straightforward approaches, where the learner stores a small subset of the data for previous tasks into a memory and utilizes the memory by replaying samples to keep a model staying in a feasible region without losing the performance on the previous tasks.
Gradient episodic memory (GEM) \citep{lopez2017gradient} first formulated the replay based continual learning as a constrained optimization problem.
This formulation allows us to rephrase the constraints on objectives for previous tasks as inequalities based on the inner product of loss gradient vectors for previous tasks and a current task. 
However, the gradient update by GEM variants cannot guarantee both theoretical and empirical convergence of its constrained optimization problem.
% convergence analysis of the performance of previously learned tasks, which implies a measure of catastrophic forgetting, has not been rigorously studied in the literature.
The modified gradient updates do not always satisfy the loss constraint theoretically, and we can also observe the forgetting phenomenon occurs empirically.
It also implies that this intuitive reformulation violates the constrained optimization problem and cannot provide theoretical guarantee to prevent catastrophic forgetting without a rigorous convergence analysis.

% \textr{In this work, we provide one explanation for why catastrophic forgetting occurs by describing continual learning precisely with a smooth nonconvex finite-sum optimization problem.}
In this work, we explain the cause of catastrophic forgetting by describing continual learning with a smooth nonconvex finite-sum optimization problem.
% The nonconvex finite-sum optimization problem offers a solution to analyze catastrophic forgetting by measuring the convergence of previously learned tasks, which is related to the performance.
In the standard single task case, SGD \citep{ghadimi2013stochastic}, ADAM \citep{reddi2018convergence}, YOGI \citep{zaheer2018adaptive}, SVRG \citep{reddi2016stochastic}, and SCSG \citep{lei2017non} are the algorithms for solving nonconvex problems that arise in deep learning.
To analyze the convergence of those algorithms, previous works study the following nonconvex finite-sum problem
% Now we express our continual learning problem of the form
\begin{equation}
\label{eq:fs_problem}
    \underset{x \in \mathbb{R}^d}{\min}\ f(x)= {1 \over n} \sum_{i=1}^{n}f_i (x),
\end{equation}
where we assume that \textbf{each objective $f_i(x)$ with a model $x$ and a data point index $i\in [n]$} for a dataset with size $n$ (by the convention for notations in nonconvex optimization literature \citep{reddi2016stochastic}) is nonconvex with $L$-smoothness assumption.
In general, we denote $f_i(x)$ as $f(x ; d_i)$ where $d_i$ is a datapoint tuple \textsc{(input, output)} with index $i$.
We expect that a stochastic gradient descent based algorithm reaches a stationary point instead of the global minimum in nonconvex optimization.
Unlike the convex case, the convergence is generally measured by the expectation of the squared norm of a gradient $\mathbb{E} \lVert \nabla f(x) \rVert^2$.
The theoretical computational complexity is derived from the $\epsilon$-accurate solution, which is also known as a stationary point with $\mathbb{E} \lVert \nabla f(x) \rVert^2\leq \epsilon$.
The general nonconvex finite-sum problems assume that all data points can be sampled during training iterations.
This fact is an obstacle to directly apply (\ref{eq:fs_problem}) for continual learning problem. 
% 여기 부적절
% Suppose we divide the entire sum of objectives into two terms for previous tasks and current tasks, and measure the convergence on each term.
% Then, we can observe the transition of convergences on the previous and current tasks respectively while learning sequentially from a data stream.
% We consider this transition of convergence on the previous task as catastrophic forgetting if $\mathbb{E} \lVert \nabla f_{P}(x) \rVert^2$ with the set of data points from previous tasks $P$ increases over iterations.

We provide a solution of the above issue by leveraging memory-based methods, which allow models to access a partial access to the dataset of previous tasks.
In this setting, we can analyze nonconvex stochastic optimization problems on the convergence of previous tasks with limited access.
Similar with adaptive methods for noncovex optimization, we apply adaptive step sizes during optimization to minimize forgetting with theoretical guarantee.
Specifically, we make the following contributions:
\begin{itemize}
    \item We decompose the finite-sum problem of entire tasks into two summation terms for previous tasks and a current task, respectively. We theoretically show that small random subsets of previous tasks lead to analyzing the expected convergence rate of both tasks while learning a current task.
    \item We study the convergence of gradient methods under a small memory where the backward transfer performance degrades, and propose a new formulation of continual learning problem with the forgetting term. We then show why catastrophic forgetting occurs theoretically and empirically. 
    \item Though memory-based methods mitigate forgetting, previous works does not fully exploit the gradient information of memory. We introduce a novel adaptive method and its extension which adjust step sizes between tasks at each step with theoretical ground, and demonstrate that both methods show remarkable performance on image classification tasks. 
\end{itemize}

\section{Related Work}
\label{sec:related}
% The literature in continual learning can be divided into episodic learning and task-free learning.
% Episodic learning based methods assume that a training model is able to access clear task boundaries and stores observed examples in the task-wise replay memory \citep{lopez2017gradient,DBLP:conf/iclr/ChaudhryRRE19}.
% Based on this approach, these works can utilize the optimization problem with constraints on episodic memory.
% In the real world, an AI system experiences arbitrarily shifting data streams, which we are not able to access task boundaries in the real world.
% Task-free continual learning studies the general scenario without the task-boundary assumption in the online setting.
% \citep{aljundi2019task} introduces memory-aware synapses (MAS) and applies a learning protocol without waiting until a task is finished.
% ER-Reservoir \citep{chaudhry2019tiny} updates the fixed length memory with a probability $p$ and replay a random subset of memory.
% \citep{aljundi2019gradient} adopts the memory system of GEM selecting observed examples to store for preventing catastrophic forgetting, and has allowed us to understand task-free continual learning as a variant of constrained optimization problem.
% Unlike this prior work, we introduce a simple extension to these memory-based approaches with adaptive learning rates on both gradients from the current data stream and the memory, and provide theoretical guarantees.
%%%%%%%%%%%%%%%%%%%%%%%%%%%%%%%%%%%%%%%%%%%%%%%%%%%%%%%%%%%%%%%%%%%%%%%%%%%%%%%%%%%%

\textbf{Memory-based methods.}
% Memory-based methods store and replay a small subset of data from previous tasks.
Early memory-based methods utilize memory by the distillation \citep{rebuffi2017icarl,li2017learning} or the optimization constraint \citep{lopez2017gradient,DBLP:conf/iclr/ChaudhryRRE19}.
Especially, A-GEM \citep{DBLP:conf/iclr/ChaudhryRRE19} simplifies the approach for constraint violated update steps as the projected gradient on a reference gradient which ensures that the average memory loss over previous tasks does not increase.  
Recent works \citep{chaudhry2019tiny,chaudhry2020using,riemer2018learning} have shown that updating the gradients on memory directly, which is called experience replay, is a light and prominent approach.
% Our work differs in two important ways.
We focus on convergence of continual learning, but the above methods focus on increasing the empirical performance without theoretical guarantee.
Our analysis provides a legitimate theoretical convergence analysis under the standard smooth nonconvex finite-sum optimization problem setting.
Further, \citep{knoblauch2020optimal} shows the perfect memory for optimal continual learning is NP-hard by using set-theory, but the quantitative analysis of performance degradation is less studied.

% We extend memory based continual learning \citep{lopez2017gradient, DBLP:conf/iclr/ChaudhryRRE19,riemer2018learning,chaudhry2019tiny} to nonconvex optimization problems to provide the theoretical guarantee of previous methods.
% Our problem setting is related to the theoretical convergence analysis of smooth nonconvex optimization.

\textbf{Adaptive step sizes in nonconvex setting.} 
Adaptive step sizes under smooth nonconvex finite-sum optimization problem have been studied on general single task cases \citep{reddi2018convergence,zhang2020adaptive,zaheer2018adaptive} recently.
\citep{simsekli2019tail,zhang2020adaptive,simsekli2020fractional} have revealed that there exists a heavy-tailed noise in some optimization problems for neural networks, such as attention models, and \citep{zhang2020adaptive} shows that adaptive methods are helpful to achieve the faster convergence under the heavy-tailed distribution where stochastic gradients are poorly concentrated around the mean.
In this work, we treat the continual learning problem where stochastic gradients of previous tasks are considered as the out-of-distribution samples in regard to a current task, and develop adaptive methods which are well-performed in continual learning.

\section{Preliminaries}
\label{sec:backgrounds}
Suppose that we observe the learning procedure on a data stream of continual learning at some arbitrary observation point.
Let us consider time step $t=0$ as given observation point.
We define the previous task $\mathcal{P}$ for $t<0$ as all visited data points and the current task $\mathcal{C}$ for $t\geq 0$ as all data points which will face in the future.
Then, $P$ and $C$ can be defined as the sets of data points in $\mathcal{P}$ and $\mathcal{C}$ at time step $t=0$, respectively.
Note that the above task description is based on not a sequence of multiple tasks, but two separate sets to analyze the convergence of each of $P$ and $C$ when starting to update the given batch at the current task $\mathcal{C}$ at some arbitrary observation point.
% In this section, we will show a convergence analysis of model parameters that we have trained on $P$ is being trained for $C$.
% Thus, we simply denote a data stream of continual learning as two consecutive sets $P$ and $C$.  
We consider a continual learning problem as a smooth nonconvex finite-sum optimization problem with two decomposed objectives

\begin{align}
\label{eq:seg}
    &\underset{x \in \mathbb{R}^d}{\min}\ h(x) = {1 \over n_{f}+n_{g}} \sum_{i\in P\cup C}h_i (x),
\end{align}
where $n_f$ and $n_g$ are the numbers of elements for $P$ and $C$, and $h(x)$ can be decomposed into as follows:
\begin{align}
    % h(x)&= {n_f \over n_{f}+n_{g}}\left( {1 \over n_{f}}\sum_{i\in P}h_i (x)\right) + {n_g \over n_{f}+n_{g}} \left({1 \over n_{g}} \sum_{j\in C} h_j (x)\right) \nonumber \\
    % &= {n_f \over n_{f}+n_{g}}\left( {1 \over n_{f}}\sum_{i\in P}f_i (x)\right) + {n_g \over n_{f}+n_{g}} \left({1 \over n_{g}} \sum_{j\in C} g_j (x)\right)\nonumber \\
    h(x) &={n_f \over n_{f}+n_{g}} f(x) + {n_g \over n_{f}+n_{g}} g(x). \nonumber
\end{align}
% For clarity, we use different notations $f_i(x)$ and $g_j(x)$ for the objectives of data points $i\in P$ and $j \in C$, respectively, and $f(x)$ and $g(x)$ denotes the mean loss from $P$ and $C$.

For clarity, we use $f(x) = h(x) |_P$ and $g(x) = h(x) |_C$ for the restriction of $h$ to each dataset $P$ and $C$, respectively. 
$f_i(x)$ and $g_j(x)$ also denotes the objective terms induced from data where each index is $i \in P$ and $j \in C$, respectively.
% We additionally explain the meaning of 

%we use $g_j(x)$ and $g(x)$ as a notation of the objective for the data point $j\in C$ and the mean value over $C$, respectively.
% To ease exposition,
% For clarity, we use a different notation $g_j(x)$ for a data point $j \in C$, which is usually the same objective function for a data point $i\in P$. 

Suppose that the replay memories $M_t$ for time step $ \in [0,T]$ are random variables which are the subsets of $P\cup C$ to cover prior memory-based approaches \citep{chaudhry2019tiny,DBLP:conf/iclr/ChaudhryRRE19}.
To formulate an algorithm for memory-based approaches, we define mini-batches $I_t$ which are sampled from a memory $M_t$ at step $t$.
% Let a random variable $M_t \subset P \cup C$ be the replay memory at time step $t \in [0,T]$,
% whose union is denoted by $M:=\cup_{t} M_t.$
We now define the stochastic update of memory-based method
\begin{equation}
\label{eq:gradupdate}
    x^{t+1} = x^{t} - \alpha_{H_t} \nabla f_{I_t}(x^t) - \beta_{H_t} \nabla g_{J_t}(x^t),
\end{equation}
where $I_t \subset M_t$ and $J_t \subset C$ denote the mini-batches from the replay memory and the current data stream, respectively.
Here, $H_t$ is the union of $I_t$ and $J_t$.
% In addition, for any set $S$, $\nabla f_{S}(x^t), \nabla g_{S}(x^t)$ denote the loss gradient of a model $x^t$ at $t$ with the mini-batch $S$.
In addition, for a given set $S$, $\nabla f_{S}(x^t), \nabla g_{S}(x^t)$ denote the loss gradient of a model $x^t$ with the mini-batch $S$ at time step $t$.
The adaptive step sizes (learning rates) of $\nabla f_{I_t}(x^t)$ and $\nabla g_{J_t}(x^t)$ are denoted by $\alpha_{H_t}$ and $\beta_{H_t}$ which are the functions of $H_t$.

It should be noted the mini-batch $I_t$ from $M_t$ might contain a datapoint $j\in C$ for some cases, such as ER-Reservoir.

Throughout the paper, we assume $L$-smoothness and the following statements.
\begin{assumption}
\label{assumption:lsmooth}
$f_i$ is $L$-smooth that there exists a constant $L>0$ such that for any $x,y \in \mathbb{R}^d$,
\begin{equation}
\label{eq:lsmooth}
    \lVert \nabla f_{i}(x) - \nabla f_{i}(y) \rVert \leq L \lVert x - y \rVert
\end{equation}
where $\lVert \cdot \rVert$ denotes the Euclidean norm.
Then the following inequality directly holds that
\begin{align}
\label{eq:changelsmooth}
     -{L \over 2} \lVert x - y \rVert^{2} &\leq
     f_{i}(x) - f_{i}(y)- \langle \nabla f_{i}(y), x - y \rangle \nonumber \\
      &\leq {L \over 2} \lVert x - y \rVert^2.
\end{align}
\end{assumption}
We derive Equation \ref{eq:changelsmooth} in Appendix \ref{sec:appendproof}.
Assumption \ref{assumption:lsmooth} is a well-known and useful statement in nonconvex finite-sum optimization problem \citep{reddi2016stochastic,reddi2018convergence,zhang2020adaptive,zaheer2018adaptive}, and also helps us to describe the convergence of continual learning.
We also assume the supremum of loss gap between an initial point $x^0$ and a global optimum $x^*$ as $\Delta_f$, and the upper bound on the variance of the stochastic gradients as $\sigma_f$ in the following.
\begin{align*}
\label{eq:delta}
    \Delta_f &= \underset{x^0}{\sup} f(x^0) - f(x^*), \\   \sigma_f^2 &= \underset{x}{\sup} \ {1 \over n_f} \sum_{i=1}^{n_f} \lVert \nabla f_i(x)- \nabla f(x) \rVert^2.
\end{align*}
%%%%%%%%%%%%%%%%%%%%%%%%%%%%%%%%%%%%%%%%%%%%%%%%%%%%%%%%%%%%%%%%%%%

%%%%%%%%%%%%%%%%%%%%%%%%%%%%%%%%%%%%%%%%%%%%%%%%%%%%%%%%%%%%%%%%%%%%%%%%%%
It should be noted that $g_j(x), \nabla g_j(x)$, which denote the loss and the gradient for a current task, also satisfy all three above assumptions and the following statement.

To measure the efficiency of a stochastic gradient algorithm, we define the Incremental First-order Oracle (IFO) framework \citep{ghadimi2013stochastic}.
IFO call is defined as a unit of computational cost by taking an index $i$ which gets the pair $(\nabla f_i (x), f_i (x))$, and IFO complexity of an algorithm is defined as the summation of IFO calls during optimization.
For example, a vanilla stochastic gradient descent (SGD) algorithm requires computational  cost as much as the batch size $b_{t}$ at each step, and the IFO complexity is the sum of batch sizes $\sum_{t=1}^T b_t$.
Let $T(\epsilon)$ be the minimum number of iterations to guarantee $\epsilon$-accurate solutions.
The average bound of IFO complexity is less than or equal to $\sum_{t=1}^{T(\epsilon)} b_t = O(1/\epsilon^2)$ \citep{reddi2016stochastic}.

\begin{figure*}[t]
\centering
\includegraphics[width=0.8\linewidth]{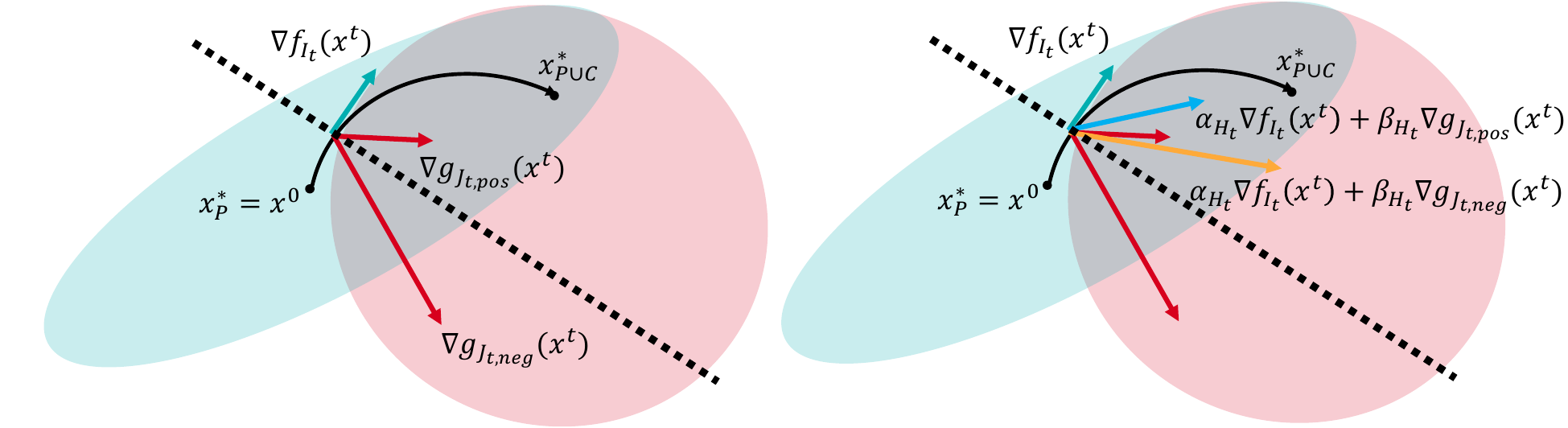}
\caption{Geometric illustration of Non-Convex Continual Learning (NCCL). In continual learning, a model parameter $x^t$ starts from a local optimal point for the previously learned tasks $x_{P}^*$. Over $T$ iterations, we expect to reach a new optimal point $x_{P\cup C}^*$ which has a good performance on both $P$ and $C$. In the $t$-th iteration, $x^t$ encounters either $\nabla g_{J_{t},pos}(x^t)$ or  $\nabla g_{J_{t},neg}(x^t)$. These two cases indicate whether $\langle f_{I_t} (x^t), \nabla g_{J_{t}}(x^t) \rangle$ is positive or not. To prevent $x^t$ from escaping the feasible region, i.e., catastrophic forgetting, we impose a theoretical condition on learning rates for $f$ and $g$.}
\label{fig:entire_procedure}
\end{figure*}

% \section{Nonconvex Continual Learning}
\section{Continual Learning as Nonconvex Optimization}
\label{sec:nccl}
We first present a theoretical convergence analysis of memory-based continual learning in nonconvex setting.
% Introduce the theoreticla analaysis
% compoensation -> update
% We use the convergence rate of stochastic gradient methods, which denotes the IFO complexity to reach an $\epsilon$-accurate solution for a smooth nonconvex finite-sum problem \textr{\citep{reddi2016stochastic}}.
% This generic form enables both deep learning and optimization communities to formulate various accelerated gradient methods with theoretical guarantee.
We aim to understand why catastrophic forgetting occurs in terms of the convergence rate,
and reformulate the optimization problem of continual learning into a nonconvex setting with theoretical guarantee.
For completeness we present all proofs in Appendix \ref{sec:appendproof}.
% plural!!! multiple algorithms: TODO
% and propose non-convex continual learning (NCCL) algorithms with theoretical convergence analysis.

%%%%%%%%%%%%%%%%%%%%%%%%%%%%%%%%%%%%%%%%%%%%%%%%%%%%%%%%%%%
\begin{algorithm}[t]
   \caption{Nonconvex Continual Learning (NCCL)}
   \label{alg:gni}
\begin{algorithmic}
   \Require Previous task set $P$, current task set $C$, initial model $x^{{0}}$.
   \State Sample a initial memory $M_0 \subset P$ \algorithmiccomment{By replay schemes, the selection dist. of $M_0$ are different.}
%   \For{$k=1$ {\bfseries to} $K$}
   %\STATE $\mathcal{U}_{f}, \; \mathcal{U}_{g} \gets \{ (x_i, y_i, f_{i}(\theta_{t_0})) \}, \{ (x_i, y_i, g_{i}(\theta_{t_0})) \}$
   \For{$t=0$ {\bfseries to} $T-1$}
   \State Sample a mini-batch $I_t \subset M_t$
   \State Sample a mini-batch $J_t \subset C$
   \State Compute step sizes $\alpha_{H_{t}}$, $\beta_{H_{t}}$ by $\nabla f_{I_t}(x^t)$, $\nabla g_{J_t}(x^{t})$
   \State $x^{t+1} \gets$  $x^t -\alpha_{H_t}\nabla f_{I_t}(x^t)- \beta_{H_{t}}\nabla g_{J_t}(x^{t})$
   \State  Update $M_{t+1}$ by the rule of replay scheme with $J_t$.
   \EndFor
%   \State $x^{0} \gets x^{T-1}$  \algorithmiccomment{The following lines are to reinitialize the observation point $t$ as $T-1$}
%   \State $P\gets P \cup \bigcup_{t} J_t$
%   \State $C \gets C - \bigcup_{t} J_t$
%   \State $M_0\gets M_{T-1}$
%   \EndFor
\end{algorithmic}
\end{algorithm}
%%%%%%%%%%%%%%%%%%%%%%%%%%%%%%%%%%%%%%%%%%%%%%%%%%%%%%%%%%%%%%%%%%

%%%%%%%%%%%%%%%%%%%%%%%%%%%%%%%%%%%%%%%%%%%%%%%%%%%%%%%%%%%%%%%%%%%%%%%%%%
% The theoretical result shows why catastrophic forgetting occurs in terms of the nonconvex optimization problem.
% As a result, we can propose the non-convex continual learning (NCCL) algorithm, where the learning rates for the previously learned tasks and the current tasks are scaled by the value of the inner product by their gradients for the parameter in Section \ref{sec:algo}.
%As discussed in Section \ref{sec:segmentation}, our objective is composed of two losses from a previously learned task and an unseen task. 
% Learning with two different balanced learning rate is our key contribution to continual learning, which balances gradients from two tasks for preventing catastrophic forgetting.
%%%%%%%%%%%%%%%%%%%%%%%%%%%%%%%%%%%%%%%%%%%%%%%%%%%%%%%%%%%%%%%%%%%%%%%%%%

\subsection{Memory-based Nonconvex Continual Learning}
\label{sec:memory_cl}

Unlike conventional smooth nonconvex finite-sum optimization problems where each mini-batch is i.i.d-sampled from the whole dataset $P\cup C$, the replay memory based continual learning encounters a non-i.i.d stream of data $C$ with access to a small sized memory $M_t$.
Algorithm \ref{alg:gni} provides the pseudocode for memory-based approach with the iterative update rule \ref{eq:gradupdate}.
% The previous task $P$ and the current task in Algorithm \ref{alg:gni} are the set of all previously visited data points and the set of data points from $K$ task streams respectively.
Now, we can analyze the convergence on $P$ and $C$ during a learning procedure on an arbitrary data stream from two consecutive sets $P$ and $C$ for continual learning \citep{DBLP:conf/iclr/ChaudhryRRE19,chaudhry2019tiny,chaudhry2020continual}.

By limited access to $P$, the expectation of gradient update $\E_{I_t \subset M_t} [\nabla f_{I_t}(x^t)]$ in Equation \ref{eq:gradupdate} for $f(x)$ is a biased estimate of the gradient $\nabla f(x^t)$.
At the timestep $t$, we have
\begin{align*}
    \nabla f_{M_t}(x^t) &=  \E_{I_t}\left[\nabla f_{I_t}(x^t) | M_t \right] = \E_{I_t}\left[\nabla f(x^t) + e_t | M_t \right] \\
    &= \nabla f(x^t) + e_{M_t},
\end{align*}
where $e_t$ and $e_{M_t}$ denote the error terms, $\nabla f_{I_t}(x^t) - \nabla f(x^t)$ and the expectation over $I_t$ given $M_t$, respectively.
It should be noted that a given replay memory $M_t$ with small size at timestep $t$ introduces an inevitable overfitting bias.

For example, there exist two popular memory schemes, episodic memory and ER-reservoir.
The episodic memory $M_t=M_0$ for all $t$ is uniformly sampled once from a random sequence of $P$, and ER-reservoir iteratively samples the replay memory $M_t$ by the selection rule $M_t \subset M_{t-1} \cup J_t.$
Here, we denote the history of $M_t$ as $M_{[0:t]}=(M_0, \cdots, M_t).$
To compute the expectation over all stochasticities of NCCL, 
we need to derive the expectation of $\nabla f_{M_t}(x^t)$ over the randomness of $M_t$.
We formalize the expectation over all learning trials with the selection randomness as follows.
% We define the selection probability as a  
% We know that 
% Unlike SGD, note that there exists an additional term on the upper bound, which is the catastrophic forgetting term $\Gamma_t$.
% In addition, the other term $B_t$ can increase the upper bound when $e_t \neq 0$.
% More specifically, in view of the expectation, $\E[B_t]$ is zero because $\E[e_t]=0$.
% This fact is based on the effect of solving the convergence on $P$ and random sampling procedure on the replay memory $M$.
\begin{lemma}
\label{lemma:memory}
If $M_0$ is uniformly sampled from $P$, 
then both episodic memory and ER-reservoir satisfies
\begin{equation*}
    \E_{M_{[0:t]}} \left[ \nabla f_{M_t} (x^t) \right] = \nabla f(x^t) \quad \text{and} \quad \E_{M_{[0:t]}}\left[ e_{M_t} \right]=0.
\end{equation*}
\end{lemma}
Note that taking expectation iteratively with respect to the history $M_{[0:t]}$ is needed to compute the expected value of gradients for $M_t$.
Surprisingly, taking the expectation of overfitting error over memory selection gets zero.
However, it does not imply $e_t=0$ for each learning trial with some $M_{[0:t]}.$

%%%%%%%%%%%%%%%%%%%%%%%%%%%%%%%%%%%%%%%%%%%%%%%%%%%%%%%%%%%%%%%%%%%%%%%%%%%%%%%%%%%%%%
\subsection{Theoretical Convergence Analysis}
\label{sec:convergence_analysis}
We now propose two terms of interest in a gradient update of nonconvex continual learning (NCCL).
% For ease of exposition,
We define
% For iteration $t\in [1,T]$ and a constant $L$,
the overfitting term $B_t$ and the catastrophic forgetting term $\Gamma_t$ as follows: %to be an expectation involving $\nabla g_{J_t}(x^t)$:
\begin{align*}
% \label{eq:cata}
    &B_t = (L\alpha_{H_t}^2 - \alpha_{H_t}) \langle \nabla f(x^t), e_t \rangle + \beta_{H_t} \langle \nabla g_{J_t}(x^t),e_t \rangle, \\
    &\Gamma_t = {\beta_{H_t}^2 L \over 2} \lVert \nabla g_{J_t}(x^t) \rVert^2 - \beta_{H_t}(1-\alpha_{H_t}L) \langle \nabla f_{I_t}(x^t), \nabla g_{J_t} (x^t) \rangle.
\end{align*}

The amount of effect on convergence by a single update can be measured by using Equation \ref{eq:changelsmooth} as follows:
\begin{align} \label{eq:expansion}
    f (x^{t+1}) &\leq f(x^t) - \langle \nabla f(x^t), \alpha_{H_t} \nabla f_{I_t}(x^t) + \beta_{H_t} \nabla g_{J_t}(x^t) \rangle \nonumber \\
    & + {L \over 2} \lVert  \alpha_{H_t} \nabla f_{I_t}(x^t) + \beta_{H_t} \nabla g_{J_t}(x^t) \rVert^2 
\end{align}
by letting $x \gets x^{t+1}$ and $y \gets x^t$.
%%%%%%%%%%%%%%%%%%%%%%%%%%%%%%%%%%%%%%%%%%%%%%%%%%%%%%%%%%
Note that the above inequality can be rewritten as
\begin{align*}
    f (x^{t+1}) &\leq  f(x^t) - \left(\alpha_{H_t} - {L \over 2} \alpha_{H_t}^2 \right) \lVert \nabla f(x^t) \rVert^2  +  \Gamma_t +  B_t \\
    &+ {L  \over 2} \alpha_{H_t}^2\lVert e_t \rVert^2.
\end{align*}

\trr{A NCCL algorithm update its model with two additional terms $B_t, \Gamma_t$ compared to conventional SGD.
An overfitting term $B_t$ and a catastrophic forgetting term $\Gamma_t$ are obtained by grouping terms that contain $e_t$ and $\nabla g_{J_t}(x^t)$, respectively.} 
These two terms inevitably degrade the performance of NCCL with respect to time.
% This reveals the basic qualitative difference between the conventional nonconvex SGD and NCCL in the convergence rate.
% Compared to nonconvex SGD methods, there exist two terms $B_t$ and $\Gamma_t$ in Equation \ref{eq:upperbound_f}.
% We group the terms containing $e_t$ into $B_t$ and the other terms into $\Gamma_t$.
It should be noted that $\Gamma_t$ has $\langle \nabla f_{I_t} (x^t), \nabla g_{J_t}(x^t) \rangle$, which is a key factor to determine interference and transfer \citep{riemer2018learning}.
% is composed of the gradients from $C$, which only exist when we update on $C$.
On the other hand, $B_t$ includes $e_t$, which is an error gradient between the batch from $M_t$ and the entire dataset $P$.
% Then, the degree of overfitting is quantified by tracing $B_t.$

Since taking the expectation over all stochasticities of NCCL implies the total expectation, we define the operator of total expectation with respect to $0\leq t <T$ for ease of exposition as follows:
\begin{equation*}
    \E_t =  \E_{M_{[0:t]}} \left[ \E_{I_t} \left[ \E_{J_t} \left[\\ \cdot \\ |I_t \right]\right] | M_{[0:t]} \right].
\end{equation*}
%%%%%%%%%%%%%%%%%%%%%%%%%%%%%%%%%%%%%%%%%%%%%%%%%%%%%%%%

% To analyze the theoretical convergence, we take the expectation with respect to the data stream $C $ and the history of memory $M_{[1:T]}.$
In addition, we denote $\E_{T-1} = \E$.
We first state the stepwise change of upper bound.
% Thm 1
% TODO: check the coefficient of B_t
\begin{lemma}\label{lemma:step}
%Suppose $f$ has $\sigma_f$ bounded gradient. $L \alpha_{H_t}^2 - \alpha_{H_t}^2 \leq \gamma$ for some $\gamma >0$ and 
Suppose that Assumption \ref{assumption:lsmooth} holds and $0 < \alpha_{H_t} \leq {2 \over L}$.
For $x^t$ updated by Algorithm \ref{alg:gni}, we have
\begin{align}
\label{eq:thm1}
    \mathbb{E}_t  \lVert \nabla f (x^t) \rVert^2 &\leq  \mathbb{E}_t \left[ { f(x^t) - f(x^{t+1}) + B_t + \Gamma_t  \over \alpha_{H_t}(1-{L\over2}\alpha_{H_t})} \right] \nonumber \\
    & \quad\quad+ \mathbb{E}_t \left[{\alpha_{H_t} L \over 2 (1-{L\over2}\alpha_{H_t})} \sigma_{f}^2 \right].
\end{align}
\end{lemma}
Surprisingly, we observe $\E_t[B_t]=0$ by Lemma \ref{lemma:memory}.
It should be also noted that the individual trial with a randomly given $M_0$ cannot cancel the effect of $B_t$.
We discuss more details of overfitting to memory in Appendix \ref{sec:overfitting}.

We now describe a convergence analysis of Algorithm \ref{alg:gni}.
We telescope over training iterations for the current task, which leads to obtain the following theorem.
\begin{theorem}
\label{thm:min}
Let $\alpha_{H_t}=\alpha={c \over \sqrt{T}} $ for some $0< c \leq {2 \sqrt{T} \over L}$ and $t\in \{0, \cdots, T-1\}$.
%$\sum \alpha_t \geq {A \over L}$ and $\alpha_t L \leq \gamma$.
By Lemma \ref{lemma:step}, the iterates of NCCL satisfy

\begin{align*}
    \underset{t}{\min}\ \mathbb{E}  \lVert \nabla f (x^t) \rVert^2  \leq {A \over \sqrt{T}} \left({1\over c}\left( \Delta_f +   \sum_{t=0}^{T-1}\E\left[ \Gamma_t \right] \right) +  {Lc \over 2} \sigma_{f}^2 \right)
\end{align*}

where $A={1 / (1- L\alpha /2 )}$. % and $\E[B_t]=0$ by taking the expectation over the random vector $M$.
\end{theorem}

% We additionally prove the convergence of $g(x)$ in Lemma \ref{lemma:g}.
% On the other hand, \trr{to ovserve the convegence of $g(x)$}.
\trr{We also prove the convergence rate of a current task $C$ with the gradient udpates from the replay-memory $M$ in continual learining.}

\begin{lemma}
\label{lemma:g}
Suppose that $I_t \cap J_t = \emptyset$,
% and the datapoints $d\in M \cap P$ use the same objective function $g_d=f_d$. 
Taking expectation over $I_t \subset M_t$ and $J_t \subset C$, we have 
\begin{equation}
     \underset{t}{\min}\ \mathbb{E}  \lVert \nabla h|_{M\cup C} (x^t) \rVert^2  \leq \sqrt{ {2 \Delta_{h|_{M\cup C}} L \over T} }\sigma_{h|_{M \cup C}},
\end{equation}
where $\Delta_{h|_{M \cup C}}$ and $\sigma_{h|_{M \cup C}}$ is the version of loss gap and the variance for $h$ on $M \cup C$, respectively.
% In fact, it should be noted that the convergence rate of $g$ is on $M\cup C$, so that it also converges to $C$ trivially.
\end{lemma}

\trr{Thus, the convergence of a current task $\gC$ is guaranteed, since its superset $M \cup C$ is converged.
Otherwise, the convergence rate might differ from the conventional SGD for $\gC$ by the given $\Delta_{ h|_{M\cup C}}, \sigma_{ h|_{M\cup C}}$ at time $0$, but the asymptotic convergence rate is still identical.}

One key observation is that $\E[\Gamma_t]$ are cumulatively added on the upper bound of $\E \lVert \nabla f(x) \rVert^2$, which is a constant in conventional SGD.
The loss gap $\Delta_f$ and the variance of gradients $\sigma_f$ are fixed values. %, so the values by dividing $\sqrt{T}$ decrease over time.
% Note that there exists a cumulative sum of $\E[\Gamma_t]$ unlike the convergence rate of SGD.
In practice, tightening $\sum_t \E[\Gamma_t]$ appears to be critical for the performance of NCCL.
However, $\sum^{T-1}_{t=0} \E[\Gamma_t] / \sqrt{T}$ is not guaranteed to converge to 0.
This fact gives rise to catastrophic forgetting in terms of a nondecreasing upper bound.
% Before showing a bound on the IFO complexity of NCCL,
\trr{We now show the key condition of the convergence of  $\sum^{T-1}_{t=0} \E[\Gamma_t] / \sqrt{T}$.} 

\begin{lemma}
\label{thm:exp_catastrophic}
    Let an upper bound $\beta > \beta_{H_t} >0$.
    % The upper bound of $\Gamma_t$
    Consider two cases, $\beta < \alpha$ and $\beta \geq \alpha$ for $\alpha$ in Theorem \ref{thm:min}.
    We have the following bound 
    \begin{align*}
        &\sum_{t=0}^{T-1} {\E[\Gamma_t] \over \sqrt{T}} < O\left( 1/T^{3/2} + 1 /T \right) \ \ \text{when} \ \beta < \alpha, \\
        &\sum_{t=0}^{T-1} {\E[\Gamma_t] \over \sqrt{T}} < O\left(\sqrt{T} + 1 /\sqrt{T} \right), \ \ \text{when} \ \beta \geq \alpha.           
    \end{align*}

    % For $\delta \leq {1\over \sqrt{T}}$, we have $O(1)$.
\end{lemma}
% Lemma \ref{lemma:g} shows a trivial result that $x^t$ converges to an $\epsilon$-accurate solution for $C$ and the supremum of $\lVert \nabla g_{J_t} (x^t) \rVert $ also decreases by the assumption of bounded variance $\sigma_g$.
% Now we can get the upper bound of $\sum \E[\Gamma_t]$ by Lemma \ref{lemma:g}, which leads to the following main result.

\trr{With the following theorem, we show that $f(x)$ can converge even if we have limited access to $P$.}

\begin{theorem}
\label{coro:smallbeta}
Let $\beta_{H_t} < \alpha = {c \over \sqrt{T}}$ for all $t$. Then we have the convergence rate
\begin{equation}
\label{eq:boundboundbound}
    \underset{t}{\min}\ \mathbb{E}  \lVert \nabla f (x^t) \rVert^2  \leq O \left( 1 \over \sqrt{T} \right).
\end{equation}
Otherwise, $f(x)$ is not guaranteed to converge when $\beta \geq \alpha$ and might diverge at the rate $O(\sqrt{T})$.    
\end{theorem}

\begin{corollary}
\label{coro:one}
% Let the expected stationary of $g(x)$ be $O({\delta \over \sqrt{T}})$ for a constant $\delta >0$ and the upper bound of learning rate for $g(x)$ be $\beta>0$.
% Under Lemma \ref{lemma:g}, the expectation of catastrophic forgetting term $\Gamma_t$ is $O(\beta^2 \sqrt{ {2 \Delta_g L \over t} }\sigma_g)$ for the worst case.
% Nonconvex continual learning by Equation (\ref{eq:gradupdate}) does not converge as the algorithm iterates for the worst case, where $ \underset{t}{\min}\ \mathbb{E} \lVert \nabla f (x^t) \rVert^2$ is $O(\beta^2\delta)$ for $1 \ll \beta^2\delta \sqrt{T}$.
% When $\beta^2\delta \leq {1 \over \sqrt{T}}$, we have
For $\beta_{H_t} < \alpha = {c \over \sqrt{T}}$ for all $t$, the IFO complexity of Algorithm \ref{alg:gni} to obtain an $\epsilon$-accurate solution is:
\begin{equation}
\label{eq:ifo}
    \text{IFO calls} =O({1 / \epsilon^2}).
\end{equation}

\end{corollary}

\trr{We build intuituions about the convergence condition of the previous tasks $\gP$ in Theorem \ref{coro:smallbeta}.
As empirically shown in stable A-GEM and stable ER-Reservoir \citep{mirzadeh2020understanding}, the condition of $\beta_{H_t}<\alpha$ theoretically implies that decaying step size is a key solution to continual learning considering when we pick any arbitrary observation points.}

\begin{remark}
To prevent catastrophic forgetting, the step size of $g(x)$, $\beta_{H_t}$ should be lower than the step size of $f(x)$, $\alpha_{H_t}$.
It should also be noted that $\E_{M_{[1:t]}}[B_t|M_0]$ is not always 0 for any $M_0$.
This implies that, from time step 0, each trial with different given $M_0$ also has the non-zero cumulative sum $\sum \E_{M_{[1:T]}}[B_t|M_0]$, which occurs overestimating bias theoretically.
% and degrades the performance on $P$. 
\end{remark}

The convergence rate with respect to the marginalization on $M_0$ in Theorem  \ref{coro:smallbeta} exactly match the usual nonconvex SGD rates.
The selection rules for $M_0$ with various memory schemes are important to reduce the variance of convergence rate with having the mean convergence rate as Equation \ref{eq:boundboundbound} among trials.
This is why memory schemes matters in continual learning in terms of variance.
\trr{Please see more details in Appendix \ref{sec:overfitting}.}
% Now we have the proper convergence rate of $f(x)$, and the model performance on $P$ can be maintained even after learning on $C$.
% It is noted that the reason of catastrophic forgetting can be explained remarkably from this nonconvex optimization perspective. 

\subsection{Reformulated Problem of Continual Learning}
\label{sec:reformulation}
% The local optimal point $x^t$ after learning on $C$ might be different from $x^0$, because $x^t$ moves towards $x_{P\cup C}^*$ as described in Figure \ref{fig:entire_procedure}.
% The above observation motivates the following formulation of continual learning to induce $\sum \E [\Gamma_t]$ to converge as $t\to \infty$ and keep $\epsilon$-accuracy of $f$ during $T$ iterations.
The previous section showed the essential factors in continual learning to observe the theoretical convergence rate.
\trr{The overfitting bias term $B_t$ has a strong dependence on the memory selection rule and can be computed exactly only if we can access the entire dataset $P$ during learning on $\gC$.
 In terms of expectation, we have shown that the effect of $B_t$ is negligible.
 We also show that its empirical effect is less important than $\Gamma_t$ in Figure \ref{fig2}.
Then we focus on the performance degradation by the catastrophic forgetting term $\Gamma_t$.
For every trial, the worst-case convergence is dependent on $\Delta_f +   \sum_{t=0}^{T-1}\E\left[ \Gamma_t \right]$ by Theorem \ref{thm:min}.
To tighten the upper bound and keep the model to be converged, we should minimize the cumulative sum of $\Gamma_t$.
}
We now reformulate the continual learning problem \ref{eq:seg} as follows.
% More specifically, the following problem is solved over $T$:
\begin{align}
\label{problem:our_problem}
    &\underset{\alpha_{H_t}, \ \beta_{H_t}}{\text{minimize}} \quad \sum_{t=0}^{T-1} \E[\Gamma_t] \nonumber \\
    & \quad \text{subject to} \quad 0 < \beta_{H_t}<  \alpha_{H_t} \leq {2 / L} \ \text{for all} \ t<T
\end{align}

It is noted that the above reformulation presents a theoretically guaranteed continual learning framework for memory-based approaches in nonconvex setting and the constraint is to guarantee the convergence of both $f(x)$ and $g(x)$.
 
%  Recall that we have the condition of two learning rates for the proof of Theorem \ref{thm:min} and Lemma \ref{lemma:g}.
%  This condition needs to be constraints of the above optimization problem.

% nonconvex continual learning in the setting of Corollary \ref{coro:one}.
% More precisely, we can now explain this method by (\ref{problem:our_problem}) with theoretical guarantee.

% \textbf{Fixed learning rates} imply that building a replay memory is the key to success in continual learning.
% It this memory setting, there is an abundant pool to sample $I_t$, which satisfies $\langle \nabla f_{I_t}(x^t), \nabla g_{J_t} (x^t) \rangle > 0$.
% This can reduce $\sum \E[\Gamma_t]$.

\section{Adaptive Methods for Continual Learning}
\label{sec:adaptive_lr}

As discussed in the above Section, we can solve a memory-based continual learning by minimizing $\sum_{t=0}^{T-1} \E[\Gamma_t]$.
Adaptive methods are variants of SGD, which automatically adjust the step size (learning rate) on a per-feature basis.
In this section, we review A-GEM in terms of adaptive methods, and also propose a new algorithm (NCCL) for achieving adaptivity in continual learning.
For brevity, we denote the inner product $\langle \nabla f_{I_t}(x^t), \nabla g_{J_t} (x^t) \rangle$ as $\Lambda_{H_t}$.
% The result of convergence analysis provides a simple continual learning framework that only adjusts two learning rates in Equation \ref{eq:gradupdate}.
% First, we review A-GEM and other methods with fixed learning rate such as GSS \citep{aljundi2019gradient} and ER-Reservoir \citep{chaudhry2019tiny} based on our framework.

\subsection{A-GEM} 
A-GEM \citep{DBLP:conf/iclr/ChaudhryRRE19} propose a surrogate of $\nabla g_{J_t} (x^t)$ as the following equation to avoid violating the constraint when the case of interference, $\Lambda_{H_t}\leq 0$:

\begin{equation*}
\label{eq:surrogate}
    \nabla g_{J_t}(x^t) - \left\langle {\nabla f_{I_t}(x^t) \over \lVert \nabla f_{I_t}(x^t) \rVert}, \nabla g_{J_t} (x^t) \right\rangle {\nabla f_{I_t}(x^t) \over \lVert \nabla f_{I_t}(x^t) \rVert}.
\end{equation*}
Let $\beta$ be the step size for $g(x)$ when the constraint is not violated.
Then we can interpret the surrogate as an adaptive learning rate $\alpha_{H_t}$, which is $\alpha (1 - {\langle \nabla f_{I_t}(x^t), \nabla g_{J_t} (x^t) \rangle \over \lVert \nabla f_{I_t}(x^t) \rVert^2})$ to cancel out the negative component of $\nabla f_{I_t}(x^t)$ on $\nabla g_{J_t}(x^t)$.

For the transfer case $\Lambda_{H_t}>0$, A-GEM use $\alpha_{H_t}=0$. 
After applying the surrogate, $\E[\Gamma_t]$ is reduced as shown in Appendix \ref{sec:derivation_algo}. %\ref{sec:derivation_algo}.
It is noted that A-GEM theoretically violates the constraints of (\ref{problem:our_problem}) to prevent catastrophic forgetting by letting $\alpha_{H_t}=0$ and does not utilize the better transfer effect.
Then, A-GEM is an adaptive method without theoretical guarantee.

\subsection{NCCL}
\label{sec:nccl_adaptive_method_1}

As discussed above, we note that $\E[\Gamma_t]$ is a quadratic polynomial of $\beta_{H_t}$.
For the interference case $ \Lambda_{H_t} \leq 0$, the minimum point of polynomial, $\beta_{H_t}^*$ has a negative value which violates the constraint $\beta_{H_t} > 0$, and $\E[\Gamma_t]$ is monotonically increasing on $\beta_{H_t} >0$.
Then, we instead adapt $\alpha_{H_t}$ to reduce the value of $\E[\Gamma_t]$ at time $t$ by adopting the scheme of A-GEM. 
The minimum of the polynomial on $\E[\Gamma_{t}]$ can be obtained when the case of transfer, $ \Lambda_{H_t} > 0$ by differentiating on $\beta_{H_t}$.
Then the minimum $\E[\Gamma_{t}^*]$ and the optimal step size $\beta_{H_t}^*$ can be obtained as
\begin{align*}
    &\beta_{H_t}^* = {(1-\alpha_{H_t} L)\Lambda_{H_t} \over L \lVert \nabla g_{J_t}(x^t) \rVert^2}, \quad \E[\Gamma_t^*] = - {(1-\alpha_{H_t} L)\Lambda_{H_t} \over 2L \lVert \nabla g_{J_t}(x^t) \rVert^2}.
\end{align*}

% A direct consequence $C_{I_t}^*<0$ implies that the optimal catastrophic forgetting surprisingly helps $f(x)$ to decrease the upper bound of stationary.
% We integrate the above result to propose a better nonconvex continual learning algorithm with a theoretical guarantee.
% More specifically, we propose an adaptive learning rate method that can reduce $\sum \E[\Gamma_t]$ in both cases of $\Lambda_{H_t} \leq 0$ and $\Lambda_{H_t} > 0$.
To satisfy the constraints of (\ref{problem:our_problem}), we should update $\nabla f_{I_t}(x^t)$ with non-zero step size and $\beta_{H_t} < \alpha_{H_t}$ for all $t$.
Then the proposed adaptive method for memory-based approaches is given by
\begin{align*}
\label{eq:alpha}
    &\alpha_{H_t}=
\begin{cases}
\alpha (1 - {\Lambda_{H_t} \over \lVert \nabla f_{I_t}(x^t) \rVert^2}), & \quad \quad\quad \quad \ \ \Lambda_{H_t} \leq 0 \\
\alpha, & \quad \quad\quad \quad \ \ \Lambda_{H_t} > 0,
\end{cases} \\
    &\beta_{H_t}=
\begin{cases}
\alpha, &  \Lambda_{H_t} \leq 0 \\
\min\left(\alpha(1-\delta),{(1-\alpha L)\Lambda_{H_t} \over L \lVert \nabla g_{J_t}(x^t) \rVert^2}\right), &  \Lambda_{H_t} > 0
\end{cases}
\end{align*}
where $\alpha=c/\sqrt{T}$ and $\delta$ is some constant $0<\delta\ll 1$.
\trr{Note that our two adaptive learning rates are a stepwise greedy perspective choice of memory-based continual learning.}

%%%%%%%%%%%%%%%%%%%%%%%%%%%%%%%%%%%%%%%%%%%%%%%%%%%%%%%% Results %%%%%%%%%%%%%%%%%%%%%%%%%%%%%%%%%%%%%%%%%%

\begin{figure*}[t]
  \centering
  \subcaptionbox{\label{fig2:a}}{\includegraphics[width=1.75in]{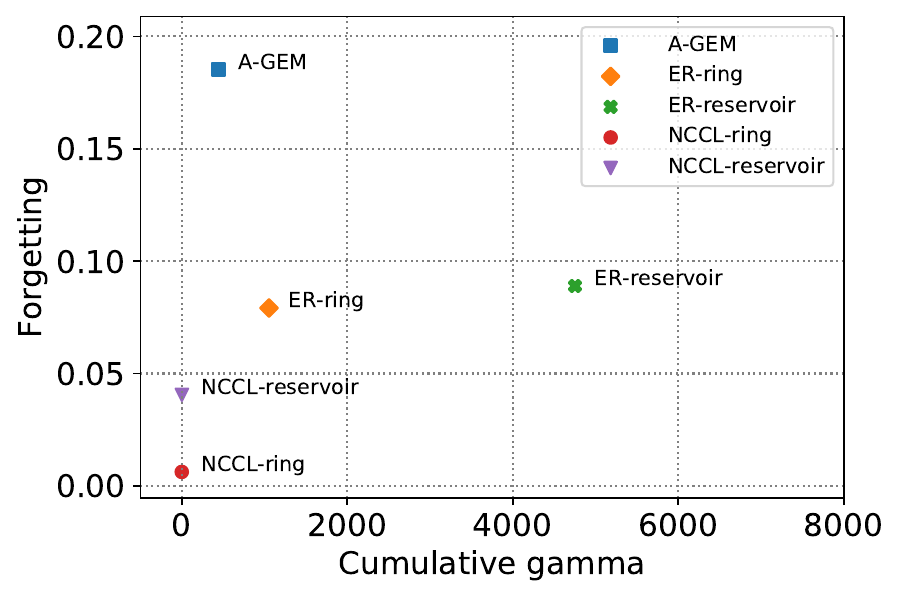}}\hspace{0.2em}%
  \subcaptionbox{\label{fig2:b}}{\includegraphics[width=1.75in]{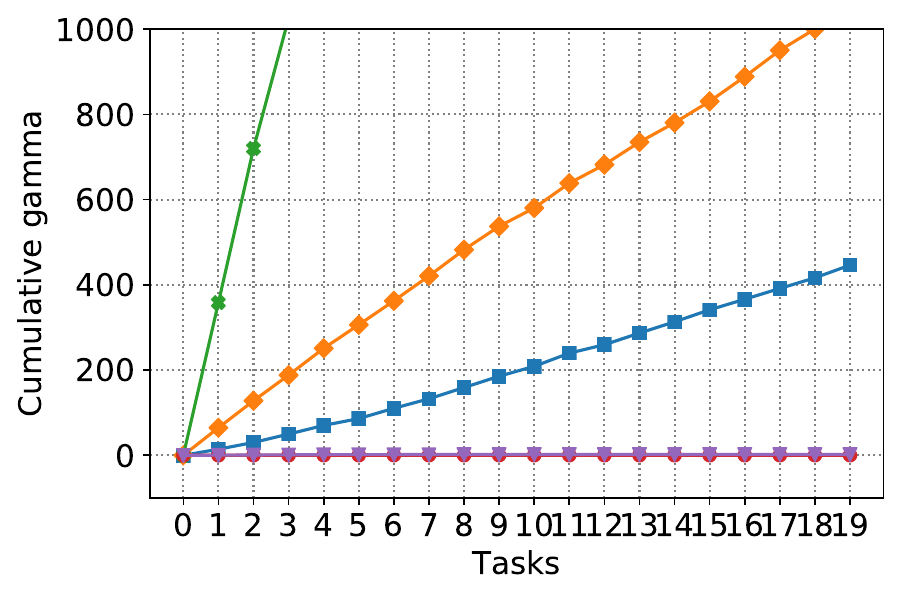}}\hspace{0.2em}%
  \subcaptionbox{\label{fig2:c}}{\includegraphics[width=1.75in]{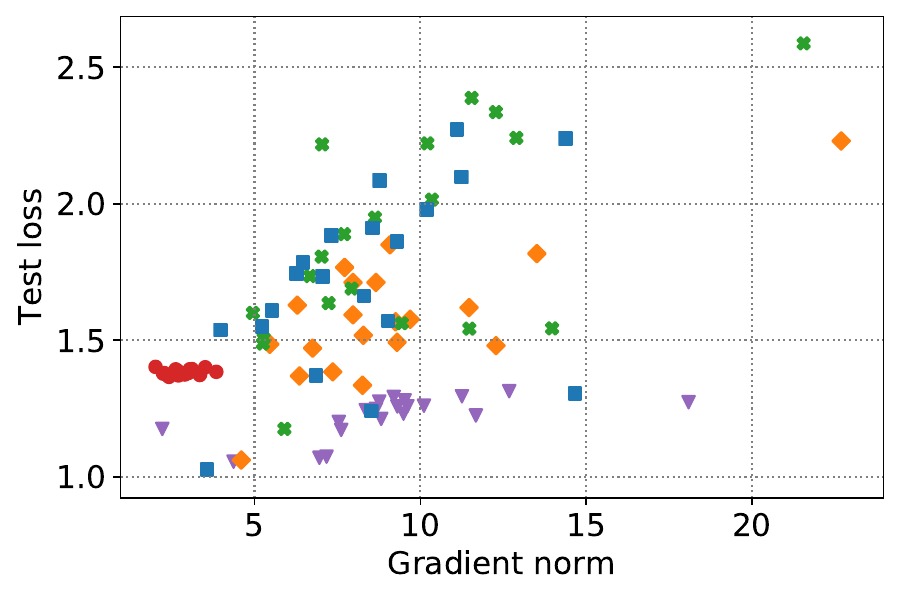}}

  \smallskip
  % Fixed length
  \centering
  \subcaptionbox{\label{fig2:d}}{\includegraphics[width=1.75in]{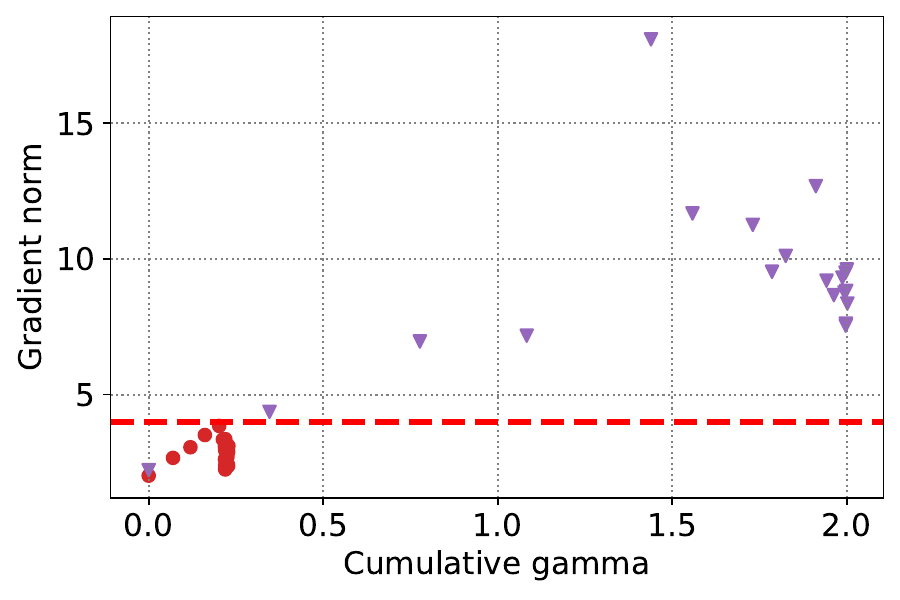}}\hspace{0.2em}%
  \subcaptionbox{\label{fig2:e}}{\includegraphics[width=1.75in]{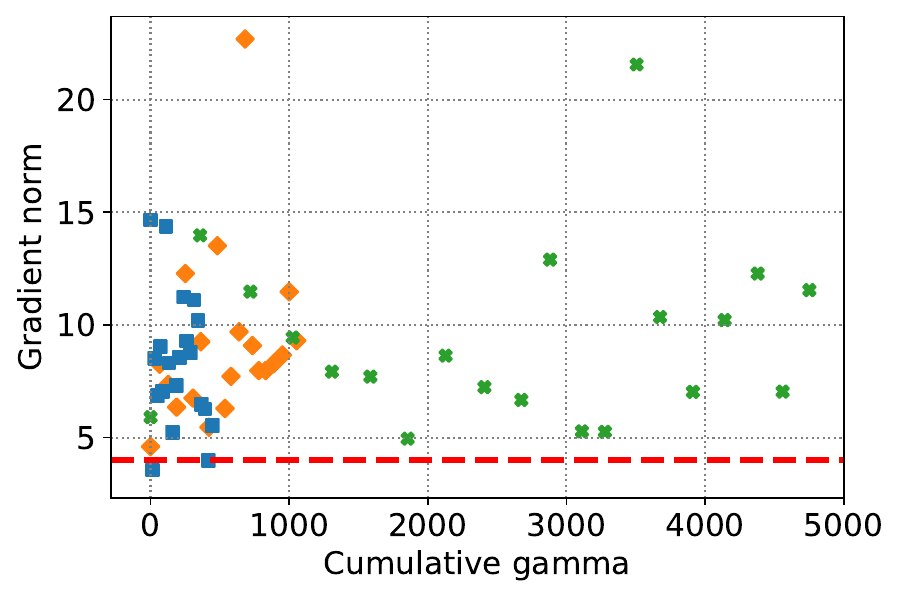}}\hspace{0.2em}%
  \subcaptionbox{\label{fig2:f}}{\includegraphics[width=1.75in]{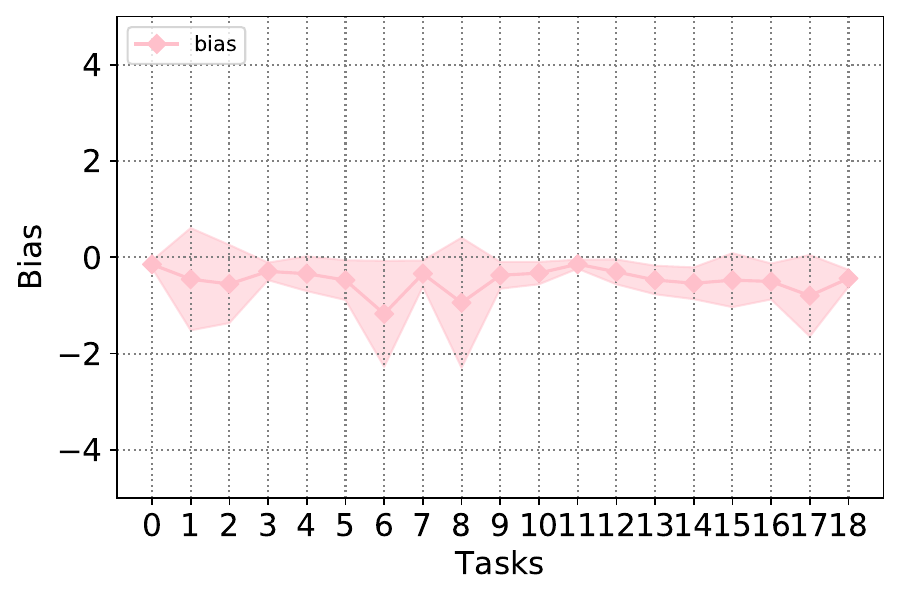}}
    \caption{Metrics for continual learning (CL) algorithms trained on split-CIFAR100 with different 5 seeds . (a) Forgetting versus $\sum \E[\Gamma_t]$ at the end of training. (b) Evolution of $\sum \E[\Gamma_t]$ during continual learning. (c) Empirical verification of the relation between $\lVert \nabla f(x) \rVert$ for the first task and test loss of the first task in split CIFAR-100. (d)-(e) are the empirical verification of $\sum \E[\Gamma_t]$ versus  $\lVert \nabla f(x) \rVert$ for the first task in CL algorithms. The red horizontal line indicates the empirical  $\lVert \nabla f(x) \rVert$ right after training the first task. (f) Illustration of empirical $B_t$ at the end of each task.}
\label{fig2}
\end{figure*}

\section{Experiments}
\label{sec:exp}

% In this section, after expaining our experimental setup, we demonstrate the performance of continual learning algorithms and proposed quantities.
We use two following metrics to evaluate algorithms. \textbf{(1) Average accuracy} is defined as ${1\over T}\sum_{j=1}^T a_{T,j}$, where $a_{i,j}$ denotes the test accuracy on task $j$ after training on task $i$.
\textbf{(2) Forgetting }is the average maximum forgetting is defined as ${1\over T-1} \sum_{j=1}^{T-1} \underset{l \in [T-1]}{\max} (a_{l,j}-a_{T,j})$.
Due to limited space, we report the details of architecture and learning procedure and missing results with additional datasets in Appendix \ref{sec:append_exp}.

\subsection{Experimental setup}
\textbf{Datasets.}
We demonstrate the experimental results on standard continual learning benckmarks:
% Now we describe the data stream for continual learning used in this section.
\textbf{Permuted-MNIST} \citep{kirkpatrick2017overcoming} is a MNIST \citep{lecun1998gradient} based dataset, where each task has a fixed permutation of pixels and transform data points by the permutation to make each task distribution unrelated.
\textbf{Split-MNIST} \citep{zenke2017continual} splits MNIST dataset into five tasks. Each task consists of two classes, for example (1, 7), (3, 4), and has approximately 12K images. 
\textbf{Split-CIFAR10, 100, and MiniImagenet} also split versions of CIFAR-10, 100 \citep{krizhevsky2009learning}, and MiniImagenet \citep{NIPS2016_90e13578} into five tasks and 20 tasks.

% Split-CIFAR10 is one of most commonly used continual learning datasets \citep{DBLP:conf/iclr/LeeHZK20,rebuffi2017icarl,zenke2017continual,lopez2017gradient,aljundi2019gradient}.

\textbf{Baselines.}
We report the experimental evaluation on the online continual setting which implies a model is trained with a single epoch.
We compare with the following continual learning baselines.
    % \textbf{iid-offline and iid-online} is a baseline model trained on an i.i.d. stream of the dataset, unlike continual learning, for a single epoch. iid-offline shows the best performance of a given model trained offline for multiple epochs.
    \textbf{Fine-tune} is a simple method that a model trains observed data naively without any support, such as replay memory.
    \textbf{Elastic weight consolidation (EWC)} is a regularization based method by Fisher Information \citep{kirkpatrick2017overcoming}.
    \textbf{ER-Reservoir} chooses samples to store from a data stream with a probability proportional to the number of observed data points. The replay memory returns a random subset of samples at each iteration for experience replay.
    ER-Reservoir \citep{chaudhry2019tiny} shows a powerful performance in continual learning scenario.
    \textbf{GEM and A-GEM} \citep{lopez2017gradient,DBLP:conf/iclr/ChaudhryRRE19} use gradient episodic memory to overcome forgetting. The key idea of GEM is gradient projection with quadratic programming and A-GEM simplifies this procedure.
    % \textbf{Gradient-based sample selection (GSS)} \citep{aljundi2019gradient} introduce a sampling method for replay memory, which maximize the diversity of gradients of data points in the memory.
    % \textbf{Continual neural Dirichlet process mixture (CN-DPM)} \citep{DBLP:conf/iclr/LeeHZK20} is a expansion-based method that allocates a subset of the data into a expert network. CN-DPM expands the number of experts under the Bayesian nonparametric framework.
    We also compare with iCarl, MER, ORTHOG-SUBSPACE \citep{chaudhry2020continual}, stable SGD \citep{mirzadeh2020understanding}, and MC-SGD \citep{mirzadeh2020linear}.

\begin{table*}[t]
    \caption{Accuary and Forgetitng results between the proposed methods (NCCL+Ring, NCCL+Reservoir) and other baselines in task-incremental learning. When the replay-memory is used, we denote the memory size as the number of examples per class per task. The additional results and the detailed setting with different memory size is in Appendix \ref{sec:append_exp}}
    \centering
    \setlength\tabcolsep{5pt}
\begin{tabular}{@{}cccccccc@{}}
\toprule
\multirow{3}{*}{\textbf{Method}} & \textbf{dataset}           & \multicolumn{2}{c}{\textbf{Permuted-MNIST}}  & \multicolumn{2}{c}{\textbf{split-CIFAR 100}} & \multicolumn{2}{c}{\textbf{split-MiniImagenet}} \\ \cmidrule(l){2-8} 
                                 & \textbf{memory size}       & \multicolumn{2}{c}{\textbf{5}}               & \multicolumn{2}{c}{\textbf{5}}               & \multicolumn{2}{c}{\textbf{1}}                  \\ \cmidrule(l){2-8} 
                                 & \textbf{memory}            & accuracy             & forgetting            & accuracy          & forgetting               & accuracy           & forgetting                 \\ \midrule
Fine-tune                        &\ding{55} & 47.9                 & 0.29(0.01)            & 40.4(2.83)        & 0.31(0.02)               & 36.1(1.31)         & 0.24(0.03)                 \\
EWC                              &\ding{55} & 63.1(1.40)           & 0.18(0.01)            & 42.7(1.89)        & 0.28(0.03)               & 34.8(2.34)         & 0.24(0.04)                 \\
stable SGD                       &\ding{55} & 80.1 (0.51)          & 0.09 (0.01)           & 59.9(1.81)        & 0.08(0.01)               & -                  & -                          \\
MC-SGD                           &\ding{55} & 85.3 (0.61)          & 0.06 (0.01)           & 63.3 (2.21)       & 0.06 (0.03)              & -                  & -                          \\
A-GEM                            &\ding{51} & 64.1(0.74)           & 0.19(0.01)            & 59.9(2.64)        & 0.10(0.02)               & 42.3(1.42)         & 0.17(0.01)                 \\
ER-Ring                          &\ding{51} & 75.8(0.24)           & 0.07(0.01)            & 62.6(1.77)        & 0.08(0.02)               & 49.8(2.92)         & 0.12(0.01)                 \\
ER-Reservoir                     &\ding{51} & 76.2(0.38)           & 0.07(0.01)            & \textbf{65.5(1.99)}        & 0.09(0.02)               & 44.4(3.22)         & 0.17(0.02)                 \\
ORHOG-subspace                   &\ding{51} & 84.32(1.1)           & 0.11(0.01)            & 64.38(0.95)       & 0.055(0.007)             & \textbf{51.4(1.44)}         & 0.10(0.01)                 \\ \midrule
NCCL + Ring                      &\ding{51} & 84.41(0.32)          & 0.053(0.002)          & 61.09(1.47)       & \textbf{0.02(0.01)}      & 45.5(0.245)        & \textbf{0.041(0.01)}       \\
NCCL+Reservoir                   &\ding{51} & \textbf{88.22(0.26)} & \textbf{0.028(0.003)} & 63.68(0.18)       & 0.028(0.009)             & 41.0(1.02)         & 0.09 (0.01)                \\ \midrule
Multi-task                         &                            & 91.3                 & 0                     & 71                & 0                        & 65.1               & 0                          \\ \bottomrule
\end{tabular}
    \label{tab:big_result}
\end{table*}

\subsection{Experiment Results}
\label{sec:expresult}
The following tables show our main experimental results, which is averaged over 5 runs.
We denote the number of examples per class per task at the top of each column.
% \end{table}
Overall, NCCL + memory schemes outperform baseline methods especially in the forgetting metric.
Our goal is to demonstrate the usefulness of the adaptive methods to reduce the catastrophic forgetting, and to show empirical evidence for our convergence analysis.
% Table \ref{tab:tab1} shows the test accuracy on three benchmark datasets. The batch size of memory and data-stream is 10 for ER-Reservoir, GSS, and our method. The memory size is 500.
We remark that NCCL successfully suppress \textbf{forgetting by a large margin} compared to baselines.
% when we compared to ER-Reservoir in particular.
It is noted that NCCL also outperforms A-GEM, which does not maximize transfer when $\Lambda_{H_t}>0$ and violates the proposed constraints in (\ref{problem:our_problem}).

We now investigate the proposed terms with regard to memory-based continual learning, $\sum \E[\Gamma_t]$ and $B_t$.
To verify our theoretical analysis, in Figure \ref{fig2} we show the cumulative catastrophic forgetting term $\sum_{t}\E[\Gamma_t]$ is the key factor of the convergence of the first task in split-CIFAR100.
During continual learning, $\sum_{t}\E[\Gamma_t]$ increases in all methods of Figure \ref{fig2:b}.
Figrure \ref{fig2:a}, \ref{fig2:d}, \ref{fig2:e} show that the larger $\sum_{t}\E[\Gamma_t]$ causes the larger forgetting and $\lVert \nabla f (x) \rVert$ for the first task. We can observe that  $\lVert \nabla f (x) \rVert$ gets larger than 4, which is for the red line, when $\sum_{t} \E[\Gamma_t]$ becomes larger than 2.
% We can observe that the empirical mean value over different selections of $M_0$ is 0 surprisingly.
% It is remarkable that that the means are negative values which are close to zero in Table \ref{tab:b}.
% In addition, we can also note that its variance is not large.}
% Therefore, the theoretical analysis on overfitting memory is verified.}
% Table \ref{tab:gamma} shows that the cumulative sum $\sum \E[\Gamma_t]$ is highly correlated with both the average test accuracy and the forgetting metric.
% It is remarkable that NCCL + Ring shows the lowest value when we compared to A-GEM, and ER-Ring.
% \textr{It is also noted that NCCL + Ring underperforms slightly on average accuracy, because the performance on new tasks is lower values by using a smaller step size for $C$ than vanilla experience replay not to violate our theoretical result.}
% Table \ref{tab:b} shows that the empirical values of mean and standard deviation, for $B_t$, over 10 random choices of $M_0$ with different seeds.
% By our theoretical results in Algorithm 1, values at each epoch are computed by all previously learned tasks $P=\cup_{k} D_k$, where the continual learning agent has seen for $k$ tasks.
We also verify that the theoretical result $\E_t[B_t]=0$ is valid in Figure \ref{fig2:f}.
It implies that the empirical results of Lemma \ref{lemma:memory}, which show the effect of $B_t$ on Equation \ref{eq:thm1}.
Furthermore, the memory bias helps to tighten the convergence rate of $P$ by having negative values in practice.
Even with tiny memory, the estimated $B_t$ has much smaller value than $\E[\Gamma_t]$ as we can observe in Figure \ref{fig2}.
For experience replay, we need not to worry about the degradation by memory bias and would like to emphasize that tiny memory can slightly help to keep the convergence on $P$ empirically.
% By considering the distribution, some of trials degrades the convergence by increasing the upper bound with the positive overfitting term, but the other trials seems to help the continual learning agent to converge on $f(x)$.
We conclude that the overfitting bias term might not be a major factor in degrading the performance of continual learning agent when it is compared to the catastrophic forgetting term $\Gamma_t$.
% The forgetting process forces the agent deviate far from the stationary point on $f(x)$, then the performance degradation by overfitting on the specific datapoints in the replay memory might be hard to occur in terms of convergence analysis. The detailed results are provided in Table \ref{tab:clipping}.
Next, we modify the clipping bound of $\beta_{H_t}$ in Section of adaptive methods to resolve the lower performance in terms of average accuracy.
In Table \ref{tab:big_result}, NCCL+Ring does not have the best average accuracy score, even though it has the lowest value of $\sum \E[\Gamma_t].$
As we discussed earlier, it is because the convergence rate of $C$ is slower than vanilla ER-Ring with the fixed step sizes.
Now, we remove the restriction of $\beta_{H_t}$,
$\min\left(\alpha(1-\delta),{(1-\alpha L)\Lambda_{H_t} \over L \lVert \nabla g_{J_t}(x^t) \rVert^2}\right)$ for $  \Lambda_{H_t} > 0$, and instead apply the maximum clipping bound $\beta_{max}$ to maximize the transfer effect, which occurs if $\Lambda_{H_t}>0$, by getting $\E[\Gamma_t^*]$.
% We test the clipped value  for Split-CIFAR 10.
% \textr{As we discussed earlier, we can prevent forgetting when $\langle \nabla f_{I_t}(x^t), \nabla g_{J_t}(x^t) \rangle>0$.
% However, we observe that $\lVert \nabla f_{I_t}(x^t) \rVert^2$ suddenly increases because of the interference at the previous step $t-1$.
% The very large learning rate $\beta_{H_t}$ by the increased $\lVert \nabla f_{I_t}(x^t) \rVert$ can force the model to fall into an arbitrary point that is likely to increase the loss of $f$.}
% It worsen the overall performance of algorithm, so we clipped $\beta_{H_t}$ to prevent this problem.
In the original version, we force $\beta_{H_t}<\alpha$ to reduce theoretical catastrophic forgetting term completely.
However, replacing with $\beta_{max}$ is helpful in terms of average accuracy as shown in Appendix \ref{sec:append_exp}.
It means that $\beta_{max}$ is a hyperparameter to increase the average accuracy by balancing between forgetting on $P$ and learning on $C$. 
In Appendix \ref{sec:append_exp}, we add more results with larger sizes of memory, which shows that NCCL outperforms in terms of average accuracy.
It means that estimating transfer and interference in terms of $\Lambda_{H_t}$ to alleviate forgetting by the small memory for NCCL is less effective.

\begin{table}[hbt!]
\caption{Results of class incremental split-CIFAR100 with Memory size = 10,000.}
\centering
\begin{tabular}{@{}cc@{}}
\toprule
\textbf{Methods}    & \textbf{accuracy} \\ \midrule
Finetune            & 3.06(0.2)       \\
A-GEM               & 2.40(0.2)       \\
GSS-Greedy \citep{aljundi2019gradient}         & 19.53(1.3)      \\
MIR \citep{DBLP:journals/corr/abs-1908-04742}               & 20.02(1.7)      \\
ER + GMED  \citep{jin2020gradient}         & 20.93(1.6) \\
MIR + GMED  \citep{jin2020gradient}         & 21.22(1.0)      \\
NCCL-Reservoir (ours)     & \textbf{21.95(0.3)}      \\ \bottomrule
\end{tabular}
\end{table}

\section{Conclusion}
\label{sec:conclusion}

We have presented a theoretical convergence analysis of continual learning.
Our proof shows that a training model can circumvent catastrophic forgetting by suppressing catastrophic forgetting term in terms of the convergence on previous task.
We demonstrate theoretically and empirically that adaptive methods with memory schemes show the better performance in terms of forgetting.
It is also noted that there exist two factors on the convergence of previous task: catastrophic forgetting and overfitting to memory.
% To tackle these problems, nonconvex continual learning uses two methods, scaling learning rates adaptive to mini-batches and sampling mini-batches with small size from the replay memory.
% Compared to other constrained optimization methods, the mechanism of NCCL utilizes both positive and negative directions between two stochastic gradients from the memory and the current task to keep a stable performance on previous tasks.
Finally, it is expected the proposed nonconvex framework is helpful to analyze the convergence rate of CL algorithms.

\begin{contributions} % will be removed in pdf for initial submission 
					  % (without ‘accepted’ option in \documentclass)
                      % so you can already fill it to test with the
                      % ‘accepted’ class option
Seungyub Han initated and led this work, implemented the proposed method, and wrote the paper.
Yeongmo Kim set up and ran baseline tests.
Taehyun Cho helped to write the paper and verified mathematical proofs.
Jungwoo Lee advised on this work.
Corresponding author: Jungwoo Lee (e-mail: junglee@snu.ac.kr)
\end{contributions}

\begin{acknowledgements} 
This work was supported in part by National Research Foundation of Korea(NRF) grants funded by the Korea Government(MSIT) (No. 2021R1A2C2014504 and No. 2021M3F3A2A02037893), in part by Institute of Information \& communications Technology Planning \& Evaluation(IITP) grants funded by the Korea Government(MSIT) (No. 2021-0-00106 (20\%), No. 2021-0-02068 (20\%), and No. 2021-0-01059 (20\%)), and in part by INMAC and BK21 FOUR program.
\end{acknowledgements}

%%%%%%%%%%%%%%%%%%%%%%%%%%%%%%%%%%%%%%%%%%%%%%%%%%%%%%%%%%%%%%%%%%%%%%%%%%%%%%%%%%%%%%%%%%%%
% End of paper
%%%%%%%%%%%%%%%%%%%%%%%%%%%%%%%%%%%%%%%%%%%%%%%%%%%%%%%%%%%%%%%%%%%%%%%%%%%%%%%%%%%%%%%%%%%%%

%%%%%%%%%%%%%%%%%%%%%%%%%%%%%%%%%%%%%%%%%%%%%%%%%%%%%%%%%%%%

% \bibliography{example_paper}
% \bibliographystyle{plain}
% Use \bibliography{yourbibfile} instead or the References section will not appear in your paper
\bibliography{han_567}

%%%%%%%%%%%%%%%%%%%%%%%%%%%%%%%%%%%%%%%%%%%%%%%%%%%%%%%%%%%%
\onecolumn %% Turn this off if single column is desired for the supplement
\appendix

%%%%%%%%%%%%%%%%%
\section{Additional Backgrounds and Extended Discussion}\label{sec:append_back}
\subsection{Summary of notations}
\begin{table}[hbt!]
\centering
\begin{tabular}{@{}cccc@{}}
\toprule
Notations     & Definitions                             & Notations                       & Definitions                           \\ \midrule
$x$           & model parameter                         & $H_t$                           & the union of $I_t$ and $J_t$          \\
$\mathcal{P}$ & previous task                          & $n_f$                           & the number of data points in $P$      \\
$\mathcal{C}$ & current task                          & $n_g$                           & the number of data points in $C$      \\
$P$           & dataset of $\mathcal{P}$                & $\langle \cdot , \cdot \rangle$ & inner product                         \\
$C$           & dataset of $\mathcal{C}$                & $L$                             & $L$-smoothness constant               \\
$h(x)$       & mean loss of $x$ on entire datasets     & $\alpha_{H_t}$  & adaptive step size for $f$ with $H_t$                       \\
$f(x)$        & mean loss of $x$ on $P$                 & $\beta_{H_t}$                   & adaptive step size for $g$ with $H_t$ \\
$g(x)$        & mean loss of $x$ on $C$                 & $M_t$                           & memory at time $t$                    \\
$f_{i}(x)$    & loss of $x$ on a data point $i\in P$    & $e_t$                           & error of estimate $f$ at time $t$     \\
$g_{j}(x)$    & loss of $x$ on a data point $j \in C$   & $e_{M_t}$                       & error of estimate $f$ with $M_t$      \\
$f_{I_t}(x)$  & mini-batch loss of $x$ on a batch $I_t$ & $f_{M_t}$                       & mean loss of $x$ with $M_t$           \\
$g_{J_t}(x)$ & mini-batch loss of $x$ on a batch $J_t$ & $M_{[t1:t2]}$   & the history of memory from $t1$ to $t2$                     \\
$I_t$         & minibatch sampled from $P$              & $B_t$                           & memory bias term at $t$           \\
$J_t$         & minibatch sampled from $C$              & $\Gamma_t$                      & forgetting term at $t$            \\
$\E_t$       & total expectation from 0 to time $t$    & $\Lambda_{H_t}$ & inner product between $\nabla f_{I_t}$ and $\nabla g_{J_t}$ \\ \bottomrule
\end{tabular}
\end{table}

\subsection{Review of terminology}
\textbf{(Restriction of $f$)} If $f : A \rightarrow B$ and if $A_0$ is a subset of $A$, then the \textbf{restriction of $f$ to $A_0$} is the function
\begin{equation*}
    f|_{A_0} : A_0 \rightarrow B
\end{equation*}
given by $f|_{A_0}(x) = f(x)$ for $x \in A_0$.

\subsection{Additional Related work}

\textbf{Regularization based methods.} EWC has an additional penalization loss that prevent the update of parameters from losing the information of previous tasks. When we update a model with EWC, we have two gradient components from the current task and the penalization loss.

\textbf{task-specific model components.} SupSup learns a separate subnetwork for each task to predict a given data by superimposing all supermasks. It is a novel method to solve catastrophic forgetting with taking advantage of neural networks.

\textbf{SGD methods without expereince replay.}  stable SGD \citep{mirzadeh2020understanding} and MC-SGD \citep{jin2021gradient} show overall higher performance in terms of average accuracy than the proposed algorithm. For average forgetting, our method has the lowest value, which means that NCCL prevents catastrophic forgetting successfully with achieving the reasonable performance on the current task. We think that our method is focused on reducing catastrophic forgetting as we defined in the reformulated continual learning problem (12), so our method shows the better performance on average forgetting. Otherwise, MC-SGD finds a low-loss paths with mode-connectivity by updating with the proposed regularization loss. This procedure implies that a continual learning model might find a better local minimum point for the new (current) task than NCCL.

For non-memory based methods, the theoretical measure to observe forgetting and convergence during training does not exist. Our theoretical results are the first attempt to analyze the convergence of previous tasks during continual learning procedure. In future work, we can approximate the value of  with fisher information for EWC and introduce Bayesian deep learning to analyze the convergence of each subnetworks for each task in the case of SupSup \citep{wortsman2020supermasks}.

\section{Additional Experimental Results and Implementation Details}\label{sec:append_exp}
 We implement the baselines and the proposed method on Tensorflow 1. For evaluation, we use an NVIDIA 2080ti GPU along with 3.60 GHz Intel i9-9900K CPU and 64 GB RAM.

\subsection{Architecture and Training detail} 
For fair comparison, we follow the commonly used model architecture and hyperparameters of \citep{DBLP:conf/iclr/LeeHZK20, chaudhry2020continual}.
For Permuted-MNIST and Split-MNIST, we use fully-connected neural networks with two hidden layers of $[400,400]$ or $[256,256]$ and ReLU activation. ResNet-18 with the number of filters $n_f=64, 20$ \citep{he2016deep} is applied for Split CIFAR-10 and 100.
 All experiments conduct a single-pass over the data stream. It is also called 1 epoch or 0.2 epoch (in the case of split tasks). We deal both cases with and without the task identifiers in the results of split-tasks to compare fairly with baselines. Batch sizes of data stream and memory are both 10.
 All reported values are the average values of 5 runs with diffrent seeds, and we also provide standard deviation. Other miscellaneous settings are the same as in \citep{chaudhry2020continual}.

\subsection{Hyperparameter grids}
We report the hyper-paramters grid we used in our experiments below.
Except for the proposed algorithm, we adopted the hyper-paramters that are reported in the original papers.
We used grid search to find the optimal parameters for each model.
\begin{itemize}
    \item finetune
    - learning rate [0.003, 0.01, 0.03 (CIFAR), 0.1 (MNIST), 0.3, 1.0]

\item EWC
    - learning rate: [0.003, 0.01, 0.03 (CIFAR),
    0.1 (MNIST), 0.3, 1.0]
    - regularization: [0.1, 1, 10 (MNIST,CIFAR), 100, 1000]

\item A-GEM
    - learning rate: [0.003, 0.01, 0.03 (CIFAR), 0.1 (MNIST), 0.3, 1.0]

\item ER-Ring
    - learning rate: [0.003, 0.01, 0.03 (CIFAR), 0.1 (MNIST), 0.3, 1.0]

\item ORTHOG-SUBSPACE
    - learning rate: [0.003, 0.01, 0.03, 0.1 (MNIST), 0.2, 0.4 (CIFAR), 1.0]

\item MER
    - learning rate: [0.003, 0.01, 0.03 (MNIST, CIFAR), 0.1, 0.3, 1.0]
    - within batch meta-learning rate: [0.01, 0.03, 0.1
    (MNIST, CIFAR), 0.3, 1.0]
    - current batch learning rate multiplier: [1, 2, 5 (CIFAR), 10 (MNIST)]

\item iid-offline and iid-online
    - learning rate [0.003, 0.01, 0.03 (CIFAR), 0.1 (MNIST), 0.3, 1.0]

\item ER-Reservoir
    - learning rate: [0.003, 0.01, 0.03, 0.1 (MNIST, CIFAR), 0.3, 1.0]

\item NCCL-Ring (default)
    - learning rate $\alpha$: [0.003, 0.001(CIFAR), 0.01, 0.03, 0.1, 0.3, 1.0]

\item NCCL-Reservoir
    - learning rate $\alpha$: [0.003(CIFAR), 0.001, 0.01, 0.03, 0.1, 0.3, 1.0]
\end{itemize}

\subsection{Hyperparameter Search on $\beta_{max}$ and Training Time}
\begin{table}[hbt!]
\caption{Permuted-MNIST (23 tasks 10000 examples per task), FC-[256,256] and Multi-headed split-CIFAR100, full size Resnet-18.  Accuracies with different clipping rate on NCCL + Ring.}
\centering
\begin{tabular}{@{}ccc@{}}
\toprule
\textbf{$\beta_{max}$} & \textbf{Permuted-MNIST} & \textbf{Split-CIFAR100} \\ \midrule
0.001                  & 72.52(0.59)             & 49.43(0.65)             \\
0.01                   & 72.93(1.38)             & 56.95(1.02)             \\
0.05                   & 72.18(0.77)             & 56.35(1.42)             \\
0.1                    & 72.29(1.34)             & 58.20(0.155)            \\
0.2                    & 74.38(0.89)             & 57.60(0.36)             \\
0.5                    & 72.95(0.50)             & 59.06(1.02)             \\
1                      & 72.92(1.07)             & 57.43(1.33)             \\
5                      & 72.31(1.79)             & 57.75(0.24)             \\ \bottomrule
\end{tabular}
\label{tab:clipping}
\end{table}

\begin{table}[hbt!]
\caption{Permuted-MNIST (23 tasks 10000 examples per task), FC-[256,256] and Multi-headed split-CIFAR100, full size Resnet-18. Training time.}
\centering
\begin{tabular}{@{}ccc@{}}
\toprule
\multirow{2}{*}{\textbf{Methods}} & \multicolumn{2}{c}{\textbf{Training time {[}s{]}}} \\ \cmidrule(l){2-3} 
                                  & \textbf{Permuted-MNIST}  & \textbf{Split-CIFAR100} \\ \midrule
fine-tune                         & 91                       & 92                      \\
EWC                               & 95                       & 159                     \\
A-GEM                             & 180                      & 760                     \\
ER-Ring                           & 109                      & 129                     \\
ER-Reservoir                      & 95                       & 113                     \\
ORTHOG-SUBSPACE                   & 90                       & 581                     \\
NCCL+Ring                         & 167                      & 248                     \\
NCCL+Reservoir                    & 168                      & 242                     \\ \bottomrule
\end{tabular}
\end{table}
\clearpage

\subsection{Additional Experiment Results}
\label{sec:addresult}

\begin{table*}[hbt!]
\caption{Permuted-MNIST (23 tasks 60000 examples per task), FC-[256,256].}
    \centering
\begin{tabular}{@{}cccccc@{}}
\toprule
\multirow{2}{*}{\textbf{Method}} & \textbf{memory size}       & \multicolumn{2}{c}{\textbf{1}} & \multicolumn{2}{c}{\textbf{5}} \\ \cmidrule(l){2-6} 
                                 & \textbf{memory}            & accuracy       & forgetting    & accuracy      & forgetting     \\ \midrule
multi-task                       & \ding{55} & 83             & -             & 83            & -              \\
Fine-tune                        & \ding{55} & 53.5 (1.46)     & 0.29 (0.01)    & 47.9          & 0.29 (0.01)     \\
EWC                              & \ding{55} & 63.1 (1.40)     & 0.18 (0.01)    & 63.1 (1.40)    & 0.18 (0.01)     \\
stable SGD                       & \ding{55} & 80.1 (0.51)     & 0.09 (0.01)    & 80.1 (0.51)    & 0.09 (0.01)     \\
MC-SGD                         & \ding{55} & 85.3 (0.61)     & 0.06 (0.01)    & 85.3 (0.61)    & 0.06 (0.01)     \\
MER                              & \ding{51} & 69.9 (0.40)     & 0.14 (0.01)    & 78.3 (0.19)    & 0.06 (0.01)     \\
A-GEM                            & \ding{51} & 62.1 (1.39)     & 0.21 (0.01)    & 64.1 (0.74)    & 0.19 (0.01)     \\
ER-Ring                          & \ding{51} & 70.2 (0.56)     & 0.12 (0.01)    & 75.8 (0.24)    & 0.07 (0.01)     \\
ER-Reservoir                     & \ding{51} & 68.9 (0.89)     & 0.15 (0.01)    & 76.2 (0.38)    & 0.07 (0.01)     \\
ORHOG-subspace                   & \ding{51} & 84.32 (1.10)    & 0.12 (0.01)    & 84.32 (1.1)    & 0.11 (0.01)     \\ \midrule
NCCL + Ring                      & \ding{51} & 74.22 (0.75)    & 0.13 (0.007)   & 84.41 (0.32)   & 0.053 (0.002)   \\
NCCL+Reservoir                   & \ding{51} & 79.36 (0.73)    & \textbf{0.12 (0.007)}   & \textbf{88.22 (0.26)}   & \textbf{0.028 (0.003)}   \\ \bottomrule
\end{tabular}
    \label{tab:permuted_60000}
\end{table*}

\begin{table*}[hbt!]
\caption{Multi-headed split-CIFAR100, reduced size Resnet-18 $n_f=20$.}
\centering
\begin{tabular}{@{}cccccc@{}}
\toprule
\multirow{2}{*}{\textbf{Method}} & \textbf{memory size}       & \multicolumn{2}{c}{\textbf{1}} & \multicolumn{2}{c}{\textbf{5}} \\ \cmidrule(l){2-6} 
                                 & \textbf{memory}            & accuracy       & forgetting    & accuracy      & forgetting     \\ \midrule
EWC                              & \ding{55} & 42.7 (1.89)     & 0.28 (0.03)    & 42.7 (1.89)    & 0.28 (0.03)     \\
Fintune                          & \ding{55} & 40.4 (2.83)     & 0.31 (0.02)    & 40.4 (2.83)    & 0.31 (0.02)     \\
Stable SGD                       & \ding{55} & 59.9 (1.81)     & 0.08 (0.01)    & 59.9 (1.81)    & 0.08 (0.01)     \\
MC-SGD                       & \ding{55} & 63.3 (2.21)     & 0.06 (0.03)    & 63.3 (2.21)    & 0.06 (0.03)     \\
A-GEM                            & \ding{51} & 50.7 (2.32)     & 0.19 (0.04)    & 59.9 (2.64)    & 0.10 (0.02)     \\
ER-Ring                          & \ding{51} & 56.2 (1.93)     & 0.13 (0.01)    & 62.6 (1.77)    & 0.08 (0.02)     \\
ER-Reservoir                     & \ding{51} & 46.9 (0.76)     & 0.21 (0.03)    & 65.5 (1.99)    & 0.09 (0.02)     \\
ORTHOG-subspace                  & \ding{51} & 58.81 (1.88)    & 0.12 (0.02)    & 64.38 (0.95)   & 0.055 (0.007)   \\ \midrule
NCCL + Ring                      & \ding{51} & 54.63 (0.65)    & \textbf{0.059 (0.01)}   & 61.09 (1.47)   & \textbf{0.02 (0.01)}     \\
NCCL + Reservoir                 & \ding{51} & 52.18 (0.48)    & 0.118 (0.01)   & 63.68 (0.18)   & 0.028 (0.009)   \\ \bottomrule
\end{tabular}
    \label{tab:cifar100_reduced}
\end{table*}

\begin{table}[hbt!]
\caption{Multi-headed split-MiniImagenet, full size Resnet-18 $n_f=64$. Accuracy and forgetting results.}
\centering
\begin{tabular}{@{}cccc@{}}
\toprule
\multirow{2}{*}{Method} & memory size                & \multicolumn{2}{c}{1}   \\ \cmidrule(l){2-4} 
                        & memory                     & accuracy   & forgetting \\ \midrule
Fintune                 & \ding{55} & 36.1(1.31) & 0.24(0.03) \\
EWC                     & \ding{55} & 34.8(2.34) & 0.24(0.04) \\
A-GEM                   & \ding{51} & 42.3(1.42) & 0.17(0.01) \\
MER                     & \ding{51} & 45.5(1.49) & 0.15(0.01) \\
ER-Ring                 & \ding{51} & 49.8(2.92) & 0.12(0.01) \\
ER-Reservoir            & \ding{51} & 44.4(3.22) & 0.17(0.02) \\
ORTHOG-subspace         & \ding{51} & 51.4(1.44) & 0.10(0.01) \\
NCCL + Ring & \ding{51} & 45.5(0.245) & \textbf{0.041(0.01)} \\
NCCL + Reservoir & \ding{51} & 41.0(1.02) & \textbf{0.09(0.01)} \\\bottomrule
\end{tabular}
\end{table}

\begin{table}[hbt!]
\caption{Multi-headed split-CIFAR100, full size Resnet-18 $n_f=64$. Accuracy and forgetting results.}
\centering
\begin{tabular}{@{}cccccc@{}}
\toprule
\multirow{2}{*}{\textbf{Method}} & \textbf{memory size}       & \multicolumn{2}{c}{\textbf{1}} & \multicolumn{2}{c}{\textbf{5}} \\ \cmidrule(l){2-6} 
                                 & \textbf{memory}            & accuracy       & forgetting    & accuracy      & forgetting     \\ \midrule
Fintune                          & \ding{55} & 42.6 (2.72)    & 0.27 (0.02)    & 42.6 (2.72)    & 0.27 (0.02)     \\
EWC                              & \ding{55} & 43.2 (2.77)     & 0.26 (0.02)    & 43.2 (2.77)    & 0.26 (0.02)     \\
ICRAL                            & \ding{51} & 46.4 (1.21)     & 0.16 (0.01)    & -             & -              \\
A-GEM                            & \ding{51} & 51.3 (3.49)     & 0.18 (0.03)    & 60.9 (2.5)     & 0.11 (0.01)     \\
MER                              & \ding{51} & 49.7 (2.97)     & 0.19 (0.03)    & -             & -              \\
ER-Ring                          & \ding{51} & 59.6 (1.19)     & 0.14 (0.01)    & 67.2 (1.72)    & 0.06 (0.01)     \\
ER-Reservoir                     & \ding{51} & 51.5 (2.15)     & 0.14 (0.09)    & 62.68 (0.91)   & 0.06 (0.01)     \\
ORTHOG-subspace                  & \ding{51} & 64.3 (0.59)     & 0.07 (0.01)    & 67.3 (0.98)    & 0.05 (0.01)     \\ \midrule
NCCL + Ring                      & \ding{51} & 59.06 (1.02)    & 0.03 (0.02)    & 66.58 (0.12)   & 0.004 (0.003)   \\
NCCL + Reservoir                 & \ding{51} & 54.7 (0.91)     & 0.083 (0.01)   & 66.37 (0.19)   & 0.004 (0.001)   \\ \bottomrule
\end{tabular}
\end{table}

% \begin{table}[hbt!]
% \caption{permuted-MNIST (23 tasks 10000 examples per task), FC-[256,256]. Accuracy and forgetting results.}
% \centering
% \begin{tabular}{|c|c|c|c|c|c|}
% \hline
%               & memory & \multicolumn{2}{c|}{1}    & \multicolumn{2}{c|}{5}     \\ \hline
% Method         &        & accuracy    & forgetting  & accuracy    & forgetting   \\ \hline
% multi-task     & x      & 91.3        & -           & 83          & -            \\ \hline
% Fine-tune      & x      & 50.6(2.57)  & 0.29(0.01)  & 47.9        & 0.29(0.01)   \\ \hline
% EWC            & x      & 68.4(0.76)  & 0.18(0.01)  & 63.1(1.40)  & 0.18(0.01)   \\ \hline
% MER            & o      & 78.6(0.84)  & 0.15(0.01)  & 88.34(0.26) & 0.049(0.003) \\ \hline
% A-GEM          & o      & 78.3(0.42)  & 0.21(0.01)  & 64.1(0.74)  & 0.19(0.01)   \\ \hline
% ER-Ring        & o      & 79.5(0.31)  & 0.12(0.01)  & 75.8(0.24)  & 0.07(0.01)   \\ \hline
% ER-Reservoir   & o      & 68.9(0.89)  & 0.15(0.01)  & 76.2(0.38)  & 0.07(0.01)   \\ \hline
% ORHOG-subspace & o      & 86.6(0.91)  & 0.04(0.01)  & 87.04(0.43) & 0.04(0.003)  \\ \hline
% NCCL + Ring    & o      & 74.38(0.89) & 0.05(0.009) & 83.76(0.21) & 0.014(0.001) \\ \hline
% NCCL+Reservoir & o      & 76.48(0.29) & 0.1(0.002)  & 86.02(0.06) & 0.013(0.002) \\ \hline
% \end{tabular}
% \end{table}

\begin{table}[hbt!]
\caption{permuted-MNIST (23 tasks 10000 examples per task), FC-[256,256]. Accuracy and forgetting results.}
\centering
\begin{tabular}{@{}cccccc@{}}
\toprule
\multirow{2}{*}{\textbf{Method}} & \textbf{memory size}       & \multicolumn{2}{c}{\textbf{1}} & \multicolumn{2}{c}{\textbf{5}} \\ \cmidrule(l){2-6} 
                                 & \textbf{memory}            & accuracy       & forgetting    & accuracy      & forgetting     \\ \midrule
multi-task                       & \ding{55} & 91.3           & -             & 83            & -              \\
Fine-tune                        & \ding{55} & 50.6 (2.57)     & 0.29 (0.01)    & 47.9          & 0.29 (0.01)     \\
EWC                              & \ding{55} & 68.4 (0.76)     & 0.18 (0.01)    & 63.1 (1.40)    & 0.18 (0.01)     \\
MER                              & \ding{51} & 78.6 (0.84)     & 0.15 (0.01)    & 88.34 (0.26)   & 0.049 (0.003)   \\
A-GEM                            & \ding{51} & 78.3 (0.42)     & 0.21 (0.01)    & 64.1 (0.74)    & 0.19 (0.01)     \\
ER-Ring                          & \ding{51} & 79.5 (0.31)     & 0.12 (0.01)    & 75.8 (0.24)    & 0.07 (0.01)     \\
ER-Reservoir                     & \ding{51} & 68.9 (0.89)     & 0.15 (0.01)    & 76.2 (0.38)    & 0.07 (0.01)     \\
ORHOG-subspace                   & \ding{51} & 86.6 (0.91)     & 0.04 (0.01)    & 87.04 (0.43)   & 0.04 (0.003)    \\ \midrule
NCCL + Ring                      & \ding{51} & 74.38 (0.89)    & 0.05 (0.009)   & 83.76 (0.21)   & 0.014 (0.001)   \\
NCCL+Reservoir                   & \ding{51} & 76.48 (0.29)    & 0.1 (0.002)    & 86.02 (0.06)   & 0.013 (0.002)   \\ \bottomrule
\end{tabular}
\end{table}

% \begin{table}[hbt!]
% \caption{Single-headed split-MNIST, FC-[256,256]. Accuracy and forgetting results.}
% \centering
% \resizebox{0.95\linewidth}{!}{
% \begin{tabular}{|c|c|c|c|c|c|c|c|}
% \hline
%           & memory & \multicolumn{2}{c|}{1}   & \multicolumn{2}{c|}{5} & \multicolumn{2}{c|}{50} \\ \hline
% Method     &        & accuracy    & forgetting & accuracy  & forgetting & accuracy  & forgetting  \\ \hline
% multi-task & x      & 95.2        & -          & -         & -          & -         & -           \\ \hline
% Fine-tune  & x      & 52.52(5.24) & 0.41(0.06) & -         & -          & -         & -           \\ \hline
% EWC        & x      & 56.48(6.46) & 0.31(0.05) & -         & -          & -         & -           \\ \hline
% A-GEM          & o & 34.04(7.10) & 0.23(0.11)   & 33.57(6.32) & 0.18(0.03)  & 33.35(4.52) & 0.12(0.04)    \\ \hline
% ER-Reservoir   & o & 34.63(6.03) & 0.79(0.07)   & 63.60(3.11) & 0.42(0.05)  & 86.17(0.99) & 0.13(0.016)   \\ \hline
% NCCL + Ring    & o & 34.64(3.27) & 0.55(0.03)   & 61.02(6.21) & 0.207(0.07) & 81.35(8.24) & -0.03(0.1)    \\ \hline
% NCCL+Reservoir & o & 37.02(0.34) & 0.509(0.009) & 65.4(0.7)   & 0.16(0.006) & 88.9(0.28)  & -0.125(0.004) \\ \hline
% \end{tabular}}
% \end{table}

\begin{table}[hbt!]
\caption{Single-headed split-MNIST, FC-[256,256]. Accuracy and forgetting results.}
\centering
\resizebox{\linewidth}{!}{
\begin{tabular}{@{}cccccccc@{}}
\toprule
\multirow{2}{*}{\textbf{Method}} & \textbf{memory size}       & \multicolumn{2}{c}{\textbf{1}} & \multicolumn{2}{c}{\textbf{5}} & \multicolumn{2}{c}{\textbf{50}} \\ \cmidrule(l){2-8} 
                                 & \textbf{memory}            & accuracy      & forgetting     & accuracy       & forgetting    & accuracy      & forgetting      \\ \midrule
multi-task                       & \ding{55} & 95.2          & -              & -              & -             & -             & -               \\
Fine-tune                        & \ding{55} & 52.52 (5.24)   & 0.41 (0.06)     & -              & -             & -             & -               \\
EWC                              & \ding{55} & 56.48 (6.46)   & 0.31 (0.05)     & -              & -             & -             & -               \\
A-GEM                            & \ding{51} & 34.04 (7.10)   & 0.23 (0.11)     & 33.57 (6.32)    & 0.18 (0.03)    & 33.35 (4.52)   & 0.12 (0.04)      \\
ER-Reservoir                     & \ding{51} & 34.63 (6.03)   & 0.79 (0.07)     & 63.60 (3.11)    & 0.42 (0.05)    & 86.17 (0.99)   & 0.13 (0.016)     \\ \midrule
NCCL + Ring                      & \ding{51} & 34.64 (3.27)   & 0.55 (0.03)     & 61.02 (6.21)    & 0.207 (0.07)   & 81.35 (8.24)   & -0.03 (0.1)     \\
NCCL+Reservoir                   & \ding{51} & 37.02 (0.34)   & 0.509 (0.009)   & 65.4 (0.7)      & 0.16 (0.006)   & 88.9 (0.28)    & -0.125 (0.004)  \\ \bottomrule
\end{tabular}}
\end{table}

% \begin{table}[hbt!]
% \caption{Single-headed split-MNIST, FC-[400,400] and mem. size=500(50 / cls.). Accuracy and forgetting results.}
% \centering
% \begin{tabular}{|c|c|}
% \hline
% mem=50           &             \\ \hline
% Method           & accuracy    \\ \hline
% multi-task       & 96.18       \\ \hline
% Fine-tune        & 50.9(5.53)  \\ \hline
% EWC              & 55.40(6.29) \\ \hline
% A-GEM            & 26.49(5.62) \\ \hline
% ER-Reservoir     & 85.1(1.02)  \\ \hline
% CN-DPM           & 93.23       \\ \hline
% Gdumb     & 91.9(0.5)   \\ \hline
% NCCL + Reservoir & 95.15(0.91) \\ \hline
% \end{tabular}
% \end{table}

\begin{table}[hbt!]
\caption{Single-headed split-MNIST, FC-[400,400] and mem. size=500(50 / cls.). Accuracy and forgetting results.}
\centering
\begin{tabular}{@{}cc@{}}
\toprule
\textbf{Method}  & \textbf{accuracy} \\ \midrule
multi-task       & 96.18             \\
Fine-tune        & 50.9 (5.53)        \\
EWC              & 55.40 (6.29)       \\
A-GEM            & 26.49 (5.62)       \\
ER-Reservoir     & 85.1 (1.02)        \\
CN-DPM           & 93.23             \\
Gdumb            & 91.9 (0.5)         \\
NCCL + Reservoir & 95.15 (0.91)       \\ \bottomrule
\end{tabular}
\end{table}

% \begin{table}[hbt!]
% \caption{Single-headed split-CIFAR10, full size Resnet-18
%  and mem. size=500(50 / cls.). Accuracy and forgetting results.}
% \centering
% \begin{tabular}{|c|c|}
% \hline
% mem=50             &             \\ \hline
% Method             & accuracy    \\ \hline
% iid-offline        & 93.17       \\ \hline
% iid-online         & 36.65       \\ \hline
% Fine-tune          & 12.68       \\ \hline
% EWC                & 53.49(0.72) \\ \hline
% A-GEM              & 54.28(3.48) \\ \hline
% GSS                & 33.56       \\ \hline
% Reservoir Sampling & 37.09       \\ \hline
% CN-DPM             & 41.78       \\ \hline
% NCCL + Ring        & 54.63(0.76) \\ \hline
% NCCL + Reservoir   & 55.43(0.32) \\ \hline
% \end{tabular}
% \end{table}

\begin{table}[hbt!]
\caption{Single-headed split-CIFAR10, full size Resnet-18
 and mem. size=500(50 / cls.). Accuracy and forgetting results.}
\centering
\begin{tabular}{@{}cc@{}}
\toprule
\textbf{Method}    & \textbf{accuracy} \\ \midrule
iid-offline        & 93.17             \\
iid-online         & 36.65             \\
Fine-tune          & 12.68             \\
EWC                & 53.49 (0.72)       \\
A-GEM              & 54.28 (3.48)       \\
GSS                & 33.56             \\
Reservoir Sampling & 37.09             \\
CN-DPM             & 41.78             \\ \midrule
NCCL + Ring        & 54.63 (0.76)       \\
NCCL + Reservoir   & 55.43 (0.32)       \\ \bottomrule
\end{tabular}
\end{table}

\begin{table}[hbt!]
\caption{Single-headed split-CIFAR100, Resnet18 with $n_f=20$. Memory size = 10,000. We conduct the experiment with the same setting of GMED \citep{jin2021gradient}.}
\centering
\begin{tabular}{@{}cc@{}}
\toprule
\textbf{Methods}    & \textbf{accuracy} \\ \midrule
Finetune            & 3.06(0.2)       \\
iid online          & 18.13(0.8)      \\
iid offline         & 42.00(0.9)      \\
A-GEM               & 2.40(0.2)       \\
GSS-Greedy          & 19.53(1.3)      \\
BGD                 & 3.11(0.2)       \\
ER-Reservoir        & 20.11(1.2)      \\
ER-Reservoir + GMED & 20.93(1.6)      \\
MIR                 & 20.02(1.7)      \\
MIR + GMED          & 21.22(1.0)      \\
NCCL-Reservoir      & \textbf{21.95(0.3)}      \\ \bottomrule
\end{tabular}
\end{table}

\clearpage

\section{Theoretical Analysis}\label{sec:appendproof}

In this section, we provide the proofs of the results for nonconvex continual learning.
We first start with the derivation of Equation \ref{eq:changelsmooth} in Assumption \ref{assumption:lsmooth}.

%%%%%%%%%%%%%%%%%%%%%%%%%%%%%%%%%%%%%%%
% \begin{assumption*}
% \label{assumption:lsmooth}
% $f_i$ is $L$-smooth that there exists a constant $L>0$ such that for any $x,y \in \mathbb{R}^d$,
% \begin{equation}
% \label{eq:lsmooth}
%     \lVert \nabla f_{i}(x) - \nabla f_{i}(y) \rVert \leq L \lVert x - y \rVert
% \end{equation}
% where $\lVert \cdot \rVert$ denotes the Euclidean norm.
% Then the following inequality directly holds that
% \begin{align}
% \label{eq:changelsmooth}
%      -{L \over 2} \lVert x - y \rVert^{2} &\leq
%      f_{i}(x) - f_{i}(y)- \langle \nabla f_{i}(y), x - y \rangle  \leq {L \over 2} \lVert x - y \rVert^2.
% \end{align}
% \end{assumption*}

\subsection{Assumption and Additional Lemma}
\begin{proof}[\textbf{Derivation of Equation \ref{eq:changelsmooth}}] %\quad
Recall that
\begin{equation}
    \left|  f_{i}(x) - f_{i}(y)- \langle \nabla f_{i}(y), x - y \rangle  \right| \leq {L \over 2} \lVert x - y \rVert^2.
\end{equation}
Note that $f_i$ is differentiable and nonconvex. We define a function $g(t)=f_i(y+t(x-y))$ for $t\in [0,1]$ and an objective function $f_i$.
By the fundamental theorem of calculus,
\begin{equation}
    \int_{0}^{1} g'(t)dt = f(x)-f(y).
\end{equation}
By the property, we have
\begin{align*}
    &\left|  f_{i}(x) - f_{i}(y)- \langle \nabla f_{i}(y), x - y \rangle  \right| \\
    &= \left| \int_{0}^1  \langle \nabla f_{i}(y+t(x-y)), x-y \rangle dt- \langle \nabla f_{i}(y), x - y \rangle  \right| \\
    &= \left| \int_{0}^1  \langle \nabla f_{i}(y+t(x-y)) - \nabla f_i(y), x-y \rangle dt \right|. 
\end{align*}
Using the Cauchy-Schwartz inequality,
\begin{align*}
   & \left| \int_{0}^1  \langle \nabla f_{i}(y+t(x-y)) - \nabla f_i(y), x-y \rangle dt \right| \\
   &\leq \left| \int_{0}^1  \lVert \nabla f_{i}(y+t(x-y)) - \nabla f_i(y)\rVert \cdot \lVert x-y \rVert dt \right|.
\end{align*}
Since $f_i$ satisfies Equation \ref{eq:lsmooth}, then we have
\begin{align*}
     &\left|  f_{i}(x) - f_{i}(y)- \langle \nabla f_{i}(y), x - y \rangle  \right| \\
     &\leq \left| \int_{0}^1  L \lVert y+t(x-y) - y \rVert \cdot \lVert x-y \rVert dt \right| \\
     & = L \lVert x-y \rVert^2 \left| \int_0^1 t dt \right| \\
     & = {L \over 2} \lVert x-y \rVert^2.
\end{align*}
\end{proof}

% Suppose that an objective function $f_i$ is $L$-smooth over \textbf{dom}$f$.
% Then for any $x, y \in \textbf{dom} f$
% for some constant $L$ Equation \ref{eq:lsmooth}
%%%%%%%%%%%%%%%%%%%%%%%%%%%%%%%%%%%%%%%%%

\begin{lemma}
\label{thm:inner_two_rand_vec}
    Let $p=[p_1, \cdots p_{D}], \ q=[q_1, \cdots, q_D]$ be two statistically independent random vectors with dimension $D$. Then the expectation of the inner product of two random vectors $\E[\langle p, q \rangle]$ is $\sum_{d=1}^{D} \E[p_d]\E[ q_d]$.
\end{lemma}
\begin{proof}
By the property of expectation,
\begin{align*}
    \E[\langle p, q \rangle] &= \E[\sum_{d=1}^D p_d q_d] \\
    &= \sum_{d=1}^D \E[ p_d q_d] \\
    &= \sum_{d=1}^D \E[ p_d] \E[q_d].
\end{align*}
\end{proof}

\subsection{Proof of Main Results}
We now show the main results of our work.
\begin{proof}[\textbf{Proof of Lemma \ref{lemma:memory}}]
To clarify the issue of $\E_{M_t} \left[ \E_{I_t} \left[e_t | M_t \right] \right]=0$, let us explain the details of constructing replay-memory as follows.
We have considered episodic memory and reservoir sampling in the paper.
We will first show the case of episodic memory by describing the sampling method for replay memory.
We can also derive the case of reservoir sampling by simply applying the result of episodic memory.

\textbf{Episodic memory (ring buffer).} 
We divide the entire dataset of continual learning into the previous task $P$ and the current task $C$ on the time step $t=0$. 
For the previous task $P$, the data stream of $P$ is i.i.d., and its sequence is random on every trial (episode).
The trial (episode) implies that a continual learning agent learns from an online data stream with two consecutive data sequences of $P$ and $C$.
Episodic memory takes the last data points of the given memory size $m$ by the First In First Out (FIFO) rule, and holds the entire data points until learning on $C$ is finished.
Then, we note that $M_t=M_0$ for all $t\geq 0$ and $M_0$ is uniformly sampled from the i.i.d. sequence of $P$.
By the law of total expectation, we derive $\E_{M_0 \subset P} \left[ \E_{I_t} \left[\nabla f_{I_t}(x^t) | M_0 \right] \right]$ for any $x^t, \ \forall t\geq 0$.

\begin{align*}
    \E_{M_0 \subset P} \left[ \E_{I_t} \left[\nabla f_{I_t}(x^t) | M_0 \right] \right] = \E_{M_0 \subset P} \left[ \nabla f_{M_0}(x^t) \right].
\end{align*}
It is known that $M_0$ was uniformly sampled from $P$ on each trial before training on the current task $C$.
Then, we take expectation with respect to every trial that implies the expected value over the memory distribution $M_0$.
We have
\begin{align*}
    \E_{M_0 \subset P} \left[ \nabla f_{M_0}(x^t) \right]=\nabla f(x^t)
\end{align*}
for any $x^t, \ \forall t$. We can consider $\nabla f_{M_t}(x^t)$ as a sample mean of $P$ on every trial for any $x^t, \ \forall t\geq 0$.
Although $x^t$ is constructed iteratively, the expected value of the sample mean for any $x^t$, $\E_{M_0 \subset P} \left[ \nabla f_{M_0}(x^t) \right]$ is also derived as $\nabla f(x^t)$. 

\textbf{Reservoir sampling.}
To clarify the notation for reservoir sampling first, we denote the expectation with respect to the history of replay memory $M_{[0:t]}=(M_0, \cdots, M_t)$ as
$\E_{M_{[0:t]}}$.
This is the revised version of $\E_{M_t}$.
Reservoir sampling is a trickier case than episodic memory, but $\E_{M_{[0:t]}} \left[ \E_{I_t} \left[e_t | M_t \right] \right]=0$ still holds.
Suppose that $M_0$ is full of the data points from $P$ as the episodic memory is sampled and the mini-batch size from $C$ is 1 for simplicity.
The reservoir sampling algorithm drops a data point in $M_{t-1}$ and replaces the dropped data point with a data point in the current mini-batch from $C$ with probability $p=m/n$, where $m$ is the memory size and $n$ is the number of visited data points so far.
The exact pseudo-code for reservoir sampling is described in [1].
The replacement procedure uniformly chooses the data point which will be dropped.
We can also consider the replacement procedure as follows.
The memory $M_t$ for $P$ is reduced in size 1 from $M_{t-1}$, and the replaced data point $d_C$ from $C$ contributes in terms of $\nabla g_{d_C}(x^t)$ if $d_C$ is sampled from the replay memory.
Let $M_{t-1} = [ d_1, \cdots, d_{|M_{t-1}|} ]$ where $| \cdot |$ denotes the cardinality of the memory.
The sample mean of $M_{t-1}$ is given as
\begin{equation}
    \nabla f_{M_{t-1}} (x^{t-1}) = {1 \over |M_{t-1}|} \sum_{d_i} \nabla f_{d_i} (x^{t-1}).
\end{equation}

By the rule of reservoir sampling, we assume that the replacement procedure reduces the memory from $M_{t-1}$ to $M_t$ with size $|M_{t-1}| -1$ and the set of remained upcoming data points $C_t\in C$ from the current data stream for online continual learning is reformulated into $C_{t-1} \cup [d_C]$.
Then, $d_C$ can be resampled from $C_{t-1} \cup [d_C]$ to be composed of the minibatch of reservoir sampling with the dfferent probability.
However, we ignore the probability issue now to focus on the effect of replay-memory on $\nabla f$.
Now, we sample $M_t$ from $M_{t-1}$, then we get the random vector $\nabla f_{M_{t}} (x^t)$ as
\begin{equation}
   \nabla f_{M_{t}} (x^t) =  {1 \over |M_{t}|} \sum_{j=1}^{|M_{t-1}|} W_{ij} \nabla f_{d_j} (x^t),
\end{equation}
where the index $i$ is uniformly sampled from $i \sim [1, \cdots, |M_{t-1}|]$, and $W_{ij}$ is the indicator function that $W_{ij}$ is 0 if $i=j$ else 1.

The above description implies the dropping rule, and $M_t$ can be considered as an uniformly sampled set with size $|M_t|$ from $M_{t-1}$.
There could also be $M_{t} = M_{t-1}$ with probability $1-p=1-m/n$.
Then the expectation of $\nabla f_{M_{t}} (x^t)$ given $M_{t-1}$ is derived as
\begin{align*}
    \E_{M_t}[ \nabla f_{M_{t}} (x^t) | M_{t-1}] &= p\left({1 \over |M_{t-1}|} \sum_{i}^{|M_{t-1}|} {1 \over |M_{t}|} \sum_{j=1}^{|M_{t-1}|}  W_{ij} \nabla f_{d_j} (x^t)\right) + (1-p)\left(\nabla f_{M_{t-1}} (x^t)\right) \\
    &= \nabla f_{M_{t-1}} (x^t).
\end{align*}
When we consider the mini-batch sampling, we can formally reformulate the above equation as
\begin{equation}
    \E_{M_t \sim p(M_t|M_{t-1})} \left[ \E_{I_t \subset M_t} \left[\nabla f_{I_t} (x^t) | M_t\right] | M_{t-1} \right]=\nabla f_{M_{t-1}} (x^t).
\end{equation}
Now, we apply the above equation recursively.
Then,
\begin{equation}
    \E_{M_1\sim p(M_1|M_0)}\left[ \cdots \E_{M_t \sim p(M_t|M_{t-1})} \left[ \E_{I_t \subset M_t} \left[\nabla f_{I_t} (x^t) | M_t\right] | M_{t-1} \right]\cdots|M_0 \right]=\nabla f_{M_{0}} (x^t).
\end{equation}
Similar to episodic memory, $M_0$ is uniformly sampled from $P$. Therefore, we conclude that

\begin{equation}
    \E_{M_0, \cdots, M_t}[\nabla f_{M_t} (x^t)]=\nabla f(x^t)
\end{equation}
by taking expectation over the history $M_{[0:t]}=(M_1, M_2, \cdots, M_t)$.

Note that taking expectation iteratively with respect to the history $M_{[t]}$ is needed to compute the expected value of gradients for $M_t$.
However, the result $\E_{M_0, \cdots, M_t}[\E_{I_t}[e_t|M_t]]=0$ still holds in terms of expectation.

Furthermore, we also discuss that the effect of reservoir sampling on the convergence of $C$.
Unlike we simply update $g(x)$ by the stochastic gradient descent on $C$, the datapoints $d\in M \cap C$ have a little larger sampling probability than other datapoints $d_{C-M} \in C - M$. The expectation of gradient norm on the averaged loss $\E \lVert \nabla g (x^t) \rVert^2$ is based on the uniform and equiprobable sampling over $C$, but the nature of reservoir sampling distort this measure slightly.
In this paper, we focus on the convergence of the previous task $C$ while training on the current task $C$ with several existing memory-based methods.
Therefore, analyzing the convergence of reservoir sampling method will be a future work.

\end{proof}

\begin{proof}[\textbf{Proof of Lemma \ref{lemma:step}}]
We analyze the convergence of nonconvex continual learning with replay memory here.
Recall that the gradient update is the following
\begin{align*}
    x^{t+1} = x^{t} - \alpha_{H_t} \nabla f_{I_t}(x^t) - \beta_{H_t} \nabla g_{J_t}(x^t)
\end{align*}
for all $t \in \{1,2, \cdots, T\}$.
Let $e_t = \nabla f_{I_t}(x^t) - \nabla f(x^t)$.
Since we assume that $f, \ g$ is $L$-smooth,
we have the following inequality by applying Equation \ref{eq:changelsmooth}:
\begin{align}
\label{aeq:lsmoothanal}
    f& (x^{t+1}) \leq f(x^t) + \langle \nabla f(x^t), x^{t+1} - x^t \rangle + {L \over 2} \lVert x^{t+1} - x^t \rVert^2 \nonumber \\
    &= f(x^t) - \langle \nabla f(x^t), \alpha_{H_t} \nabla f_{I_t}(x^t) + \beta_{H_t} \nabla g_{J_t}(x^t) \rangle + {L \over 2} \lVert  \alpha_{H_t} \nabla f_{I_t}(x^t) + \beta_{H_t} \nabla g_{J_t}(x^t) \rVert^2 \nonumber \\
    &= f(x^t) - \alpha_{H_t} \langle \nabla f(x^t), \nabla f_{I_t}(x^t) \rangle - \beta_{H_t} \langle \nabla f(x^t),  \nabla g_{J_t}(x^t) \rangle \nonumber \\
    & \ \ + {L \over 2}  \alpha_{H_t}^2 \lVert \nabla f_{I_t}(x^t)\rVert^2 + {L \over 2} \beta_{H_t}^2 \lVert \nabla g_{J_t}(x^t) \rVert^2 + L \alpha_{H_t}\beta_{H_t} \langle \nabla f_{I_t}(x^t), \nabla g_{J_t} (x^t) \rangle \nonumber \\
    &= f(x^t) - \alpha_{H_t} \langle \nabla f(x^t), \nabla f(x^t) \rangle - \alpha_{H_t} \langle \nabla f(x^t), e_t \rangle - \beta_{H_t} \langle \nabla f_{I_t}(x^t),  \nabla g_{J_t}(x^t) \rangle + \beta_{H_t} \langle \nabla g_{J_t} (x^t), e_t \rangle \nonumber \\
    & \ \ + {L \alpha_{H_t}^2 \over 2}  \lVert \nabla f(x^t)\rVert^2 + L \alpha_{H_t}^2 \langle \nabla f(x^t), e_t \rangle + {L \alpha_{H_t}^2 \over 2} \lVert e_t \rVert^2 + {L\beta_{H_t}^2 \over 2}  \lVert \nabla g_{J_t}(x^t) \rVert^2 + L \alpha_{H_t}\beta_{H_t} \langle \nabla f_{I_t}(x^t), \nabla g_{J_t} (x^t) \rangle \nonumber \\
    &= f(x^t) - \left(\alpha_{H_t} - {L \over 2} \alpha_{H_t}^2 \right) \lVert \nabla f(x^t) \rVert^2  +  {L \over 2} \beta_{H_t}^2 \lVert \nabla g_{J_t}(x^t) \rVert^2  - \beta_{H_t} ( 1 - \alpha_{H_t} L ) \langle \nabla f_{I_t}(x^t),  \nabla g_{J_t}(x^t) \rangle \nonumber \\
    & \ \ + \left( L \alpha_{H_t}^2 - \alpha_{H_t} \right) \langle \nabla f(x^t), e_t \rangle + \beta_{H_t} \langle \nabla g_{J_t} (x^t), e_t \rangle + {L  \over 2} \alpha_{H_t}^2\lVert e_t \rVert^2.
\end{align}

%%%%%%%%%%%%%%%%%%%%%%%%%%%%%%%%%%%%%%%%%%%%% 여기서부
To show the proposed theoretical convergence analysis of nonconvex continual learning,
we define the catastrophic forgetting term $\Gamma_t$ and the overfitting term $B_t$ as follows:
\begin{align*}
    &B_t = (L\alpha_{H_t}^2 - \alpha_{H_t}) \langle \nabla f(x^t), e_t \rangle + \beta_{H_t} \langle \nabla g_{J_t}(x^t),e_t \rangle, \\
    &\Gamma_t = {\beta_{H_t}^2 L \over 2} \lVert \nabla g_{J_t}(x^t) \rVert^2 - \beta_{H_t}(1-\alpha_{H_t}L) \langle \nabla f_{I_t}(x^t), \nabla g_{J_t} (x^t) \rangle.
\end{align*}
Then, we can rewrite Equation \ref{aeq:lsmoothanal} as
\begin{align}
    f& (x^{t+1}) \leq f(x^t) - \left(\alpha_{H_t} - {L \over 2} \alpha_{H_t}^2 \right) \lVert \nabla f(x^t) \rVert^2  +  \Gamma_t +  B_t + {L  \over 2} \alpha_{H_t}^2\lVert e_t \rVert^2.
\end{align}

% \begin{align*}
%   \Tilde{C}_t = {L \over 2} \beta_{H_t}^2 \lVert \nabla g_{J_t}(x^t) \rVert^2 -\beta_{H_t}(1 - \alpha_{H_t} L) \langle \nabla f(x^t),  \nabla g_{J_t}(x^t) \rangle, 
% \end{align*}
%  for $t\geq 1$. 
We first note that $B_t$ is dependent of the error term $e_t$ with the batch $I_t$.
In the continual learning step, an training agent cannot access $\nabla f(x^t)$, then we cannot get the exact value of $e_t$.
Furthermore, $\Gamma_t$ is dependent of the gradients $\nabla f_{I_t}(x^t), \nabla g_{I_t}(x^t)$ and the learning rates $\alpha_{H_t}, \beta_{H_t}$.

Taking expectations with respect to $I_t$ on both sides given $J_t$, we have
\begin{align*}
    \E_{I_t}\left[f(x^{t+1})\right] &\leq \E_{I_t}\left[  f(x^t) - \left(\alpha_{H_t} - {L \over 2} \alpha_{H_t}^2 \right) \lVert \nabla f(x^t) \rVert^2  +  \Gamma_t +  B_t + {L  \over 2} \alpha_{H_t}^2\lVert e_t \rVert^2 \Big| J_t \right] \\
    &\leq \E_{I_t}\left[  f(x^t) - \left(\alpha_{H_t} - {L \over 2} \alpha_{H_t}^2 \right) \lVert \nabla f(x^t) \rVert^2 + {L  \over 2} \alpha_{H_t}^2\lVert e_t \rVert^2 \right] + \E_{I_t} \left[ \Gamma_t +  B_t  \Big| J_t \right].
\end{align*}

Now, taking expectations over the whole stochasticity we obtain
\begin{align*}
    \E \left[f(x^{t+1})\right] &\leq \E\left[  f(x^t) - \left(\alpha_{H_t} - {L \over 2} \alpha_{H_t}^2 \right) \lVert \nabla f(x^t) \rVert^2  +  \Gamma_t +  B_t + {L  \over 2} \alpha_{H_t}^2\lVert e_t \rVert^2  \right].
\end{align*}
Rearranging the terms and assume that ${1 \over 1- {L\alpha_{H_t}/ 2} } > 0$, we have

\begin{align*}
    \left(\alpha_{H_t} - {L\over 2} \alpha_{H_t}^2 \right)\E \lVert \nabla f(x^t) \rVert^2 \leq\E \left[ f(x^t) - f(x^{t+1}) + \Gamma_t + B_t + {L  \over 2} \alpha_{H_t}^2\lVert e_t \rVert^2  \right]
    % &\leq f(x^t) - f(x^{t+1}) + C_t + {L\over 2}\alpha_{H_t}^2 \sigma_f^2 + (L\alpha_{H_t}^2 -\alpha_{H_t}) \E[\langle \nabla f(x^t), e_t \rangle] .
\end{align*}
and

\begin{align*}
    \E \lVert \nabla f(x^t) \rVert^2 &\leq\E \left[ {1 \over \alpha_{H_t} (1- {L \over 2} \alpha_{H_t})} \left( f(x^t) - f(x^{t+1}) + \Gamma_t + B_t \right) + {\alpha_{H_t} L  \over 2 (1- {L \over 2} \alpha_{H_t})} \lVert e_t \rVert^2  \right] \\
     &\leq\E \left[ {1 \over \alpha_{H_t} (1- {L \over 2} \alpha_{H_t})} \left( f(x^t) - f(x^{t+1}) + \Gamma_t + B_t \right) + {\alpha_{H_t} L  \over 2 (1- {L \over 2} \alpha_{H_t})} \sigma_f^2  \right].
    % &\leq f(x^t) - f(x^{t+1}) + C_t + {L\over 2}\alpha_{H_t}^2 \sigma_f^2 + (L\alpha_{H_t}^2 -\alpha_{H_t}) \E[\langle \nabla f(x^t), e_t \rangle] .
\end{align*}

%%%여기서부터

%%%%%%%%%%%%%%%%%%%%%%%%%%%%%%%%%%%%%%%%%%%%%%%%%%%%%%%%%%%%%%%%%%%%%
\begin{comment}
\begin{align}
\label{eq:thm1_with_bias}
    \E_{I_t} \lVert \nabla f(x^t) \rVert^2 &\leq {\lambda_{H_t} \over \alpha_{H_t}} \E_{I_t} \left[   f(x^t) - f(x^{t+1}) + \tilde{C}_t + \Lambda_{H_t} \langle \nabla f(x^t), e_t \rangle \ +  L \alpha_{H_t}\beta_{H_t} \langle e_t, \nabla g_{J_t} (x^t) \rangle \right] +  {{L\over 2}\alpha_{H_t} \sigma_f^2 \over {1- {L \over 2}\alpha_{H_t}}} \nonumber \\
   % &\leq {1 \over \alpha_{H_t}(1- {L\over2}\alpha_{H_t})} \left( f(x^t) - f(x^{t+1}) + C_t +\gamma \E[\langle \nabla f(x^t), e_t \rangle] \right) + {{L\over 2}\alpha_{H_t} \sigma_f^2   \over {1- {L \over 2}\alpha_{H_t}}}.
\end{align}

Noting that $\E[ e_t]=0$ under Assumption \ref{thm:unbiased} and Lemma \ref{thm:inner_two_rand_vec}, we have
\begin{align}
\label{eq:thm1_with_bias}
    \E_{I_t} \lVert \nabla f(x^t) \rVert^2 \leq {\lambda_{H_t} \over \alpha_{H_t}} \E_{I_t} \left[ f(x^t) - f(x^{t+1}) + \tilde{C}_t \right] + {{L\over 2}\alpha_{H_t} \sigma_f^2 \over {1- {L \over 2}\alpha_{H_t}}}.
\end{align}
Finally, we define 
\begin{equation}
    C_t = \E_{J_t}[\tilde{C}_t]
\end{equation} 
and take expectations over $J_t$,
\begin{align}
% \label{eq:thm1_with_bias}
    \E \lVert \nabla f(x^t) \rVert^2 \leq {\lambda_{H_t} \over \alpha_{H_t}} \left(\E \left[ f(x^t) - f(x^{t+1})\right] + C_t \right)+ {{L\over 2}\alpha_{H_t} \sigma_f^2 \over {1- {L \over 2}\alpha_{H_t}}}.
\end{align}
\end{comment}

\end{proof}

% Furthermore, the batch size $b$
%%%%%%%%%%%%%%%%%%%%%%%%%%%%%%%%%%%%%%%%%%%%%%%%%%%%%%%%%%%%%%%%%%%%%%%%%%%%%%%%%%%%%%%%%%%%%%%%%%%%%%

\medskip

\begin{proof}[\textbf{Proof of Theorem \ref{thm:min}}]
Suppose that the learning rate $\alpha_{H_t}$ is a constant $\alpha= c / \sqrt{T}$, for $c>0$, $1-{L\over 2} \alpha = {1\over A} >0$. Then, by summing Equation \ref{eq:thm1} from $t=0$ to $T-1$, we have

\begin{align}
\label{aeq:thm1_result}
    \underset{t}{\min} \ \E \lVert \nabla f(x^t) \rVert^2 &\leq {1 \over T} \sum_{t=0}^{T-1} \E \lVert \nabla f(x^t) \rVert^2 \nonumber \\
    &\leq {1 \over 1 - {L \over 2} \alpha} \left( {1 \over \alpha T} \left( f(x^0)-f(x^{T}) + \sum_{t=0}^{T-1} \left( \E\left[ B_t + \Gamma_t \right]\right) \right)  + {L \over 2} \alpha \sigma_f^2 \right)\nonumber \\
    &= {1 \over 1 - {L \over 2} \alpha} \left( {1 \over c \sqrt{T}} \left( \Delta_f + \sum_{t=0}^{T-1} \left(\E\left[ B_t + \Gamma_t \right] \right) \right)  + {Lc \over 2 \sqrt{T}} \sigma_f^2 \right)\nonumber \\
    &= {A \over  \sqrt{T}} \left( {1 \over c} \left( \Delta_f + \sum_{t=0}^{T-1} \E\left[ B_t + \Gamma_t \right] \right) + {Lc \over 2} \sigma_f^2 \right).
\end{align}

We note that a batch $I_t$ is sampled from a memory $M_t \subset M$ which is a random vector whose element is a datapoint $d \in P \cup C$.
Then, taking expectation over $I_t \subset M_t \subset P \cup C$ implies that $\E[B_t]=0$.
Therefore, we get the minimum of expected square of the norm of gradients
\begin{align*}
    \underset{t}{\min} \ \E \lVert \nabla f(x^t) \rVert^2 \leq {A \over  \sqrt{T}} \left( {1 \over c} \left( \Delta_f + \sum_{t=0}^{T-1} \E[\Gamma_t] \right)  + {Lc \over 2} \sigma_f^2 \right).
\end{align*}

\end{proof}

% we present the convergence rate for $g(x)$.

% \begin{lemma}
% \label{lemma:g}
% Suppose that $I_t \cap J_t = \emptyset$,
% % and the datapoints $d\in M \cap P$ use the same objective function $g_d=f_d$. 
% Taking expectation over $I_t \subset M_t$ and $J_t \subset C$, we have 
% \begin{equation}
%      \underset{t}{\min}\ \mathbb{E}  \lVert \nabla g (x^t) \rVert^2  \leq \sqrt{ {2 \Delta_g L \over T} }\sigma_g,
% \end{equation}
% where $\Delta_g$ and $\sigma_g$ is the version of loss gap and the variance for $g$ on $M \cup C$, respectively.
% In fact, it should be noted that the convergence rate of $g$ is on $M\cup C$, so that it also converges to $C$ trivially.
% \end{lemma}

% \begin{replemma}{lemma:g}
% Suppose that $I_t \cap J_t = \emptyset$,
% % and the datapoints $d\in M \cap P$ use the same objective function $g_d=f_d$. 
% Taking expectation over $I_t \subset M_t$ and $J_t \subset C$, we have 
% \begin{equation}
%      \underset{t}{\min}\ \mathbb{E}  \lVert \nabla h|_{M\cup C} (x^t) \rVert^2  \leq \sqrt{ {2 \Delta_{h|_{M\cup C}} L \over T} }\sigma_{h|_{M \cup C}},
% \end{equation}
% where $\Delta_{h|_{M \cup C}}$ and $\sigma_{h|_{M \cup C}}$ is the version of loss gap and the variance for $h$ on $M \cup C$, respectively.
% % In fact, it should be noted that the convergence rate of $g$ is on $M\cup C$, so that it also converges to $C$ trivially.
% \end{replemma} 

\begin{proof}[\textbf{Proof of Lemma \ref{lemma:g}}]
To simplify the proof, we assume that learning rates $\alpha_{H_t}, \beta_{H_t}$ are a same fixed value $\beta= c' / \sqrt{T}. $
The assumption is reasonable, because it is observed that the RHS of Equation \ref{eq:thm1} is not perturbed drastically by small learning rates in $0< \alpha_{H_t}, \beta_{H_t} \leq 2 / L \ll 1$.
% This assumption is actually the case of ER-Reservoir, which shows the remarkable performance.
% , which leads to the following.
Let us denote the union of $M_t$ over time $0\leq t \leq T-1$ as $M= \bigcup_{t} M_t$.
By the assumption, it is equivalent to update on $M \cup C$.
Then, the non-convex finite sum optimization is given as
\begin{equation}
        \underset{x \in \mathbb{R}^d}{\min}\ h|_{M \cup C}(x)= {1 \over n_{g}+ |M|} \sum_{i\in M \cup C} h_i (x),
\end{equation}
where
% $g_i$ is the same function as $f_i$, and \
$|M|$ is the number of elements in $M$.
This problem can be solved by a simple SGD algorithm \citep{DBLP:conf/icml/ReddiHSPS16}.
Thus, we have
\begin{equation}
\label{aeq:g_conv}
    \underset{t}{\min}\ \mathbb{E}  \lVert \nabla h|_{M \cup C} (x^t) \rVert^2  \leq {1\over T} \sum_{t=0}^T \mathbb{E}  \lVert \nabla h|_{M \cup C} (x^t) \rVert^2 \leq \sqrt{ {2 \Delta_{h|_{M \cup C}} L \over T} }\sigma_{h|_{M \cup C}}.
\end{equation}
\end{proof}

\begin{lemma}
\label{lemma:supsigma}
For any $C \subset D \subset M\cup C$, define $\omega^2_{h|_D}$ as
\begin{align*}
    {\omega}^2_{h|_{D}}=\underset{x}{\sup} \ \E_{j\in D} \lVert \nabla h_j(x^t) - \nabla h|_{M \cup C}(x^t) \rVert^2].
\end{align*}
Then, we have
\begin{align}
     \E \lVert \nabla g_{J_t}(x^t) \rVert^2 \leq \E \lVert \nabla h|_{M \cup C}(x^t) \rVert^2+  \underset{C \subset D \subset M\cup C}\sup {\omega}^2_{h|_{D}}.
\end{align}

\end{lemma}

\begin{proof}[\textbf{Proof of Lemma \ref{lemma:supsigma}}]
We arrive at the following result by Jensen's inequality

\begin{align}
\label{eq:samplevariance}
    \underset{x}{\sup} \E_{J_t \subset C} \lVert \nabla g_{J_t}(x^t) - \nabla h|_{M \cup C}(x^t) \rVert^2 
    % &\leq \underset{x}{\sup}  \E_{J_t \subset C} \left[  \lVert \nabla g_{J_t}(x^t) - \nabla h_{M \cup C}(x^t) \rVert^2 \right] \\
    &=\underset{x}{\sup}  \E_{J_t \subset C} \left[  \lVert  \E_{j\in J_t} [\nabla h_j(x^t)] - \nabla h|_{M \cup C}(x^t) \rVert^2 \right] \\
    &\leq \underset{C \subset D \subset M\cup C}\sup \underset{x}{\sup}  \E_{J_t \subset D} \left[  \lVert  \E_{j\in J_t} [\nabla h_j(x^t)] - \nabla h|_{M \cup C}(x^t) \rVert^2 \right] \\
    % & = \E \left[ \underset{x}{\sup}  \lVert \nabla \E_{j\in J_t} [h_j(x^t)] - \nabla h_{M \cup C}(x^t) \rVert^2 \right] \\
    % & \leq \underset{C \subset D \subset M\cup C}\sup \E_{J_t \subset D}\left[   \underset{x}{\sup}  \E_{j\in J_t} [\ \lVert \nabla h_j(x^t) - \nabla h|_{M \cup C}(x^t) \rVert^2] \right] \\
    & \leq \underset{C \subset D \subset M\cup C}\sup \left[   \underset{x}{\sup}  \E_{j\in D} [\ \lVert \nabla h_j(x^t) - \nabla h|_{M \cup C}(x^t) \rVert^2] \right] \\
    % & \leq \E \left[    \E_{j\in J_t} [\lVert \underset{x}{\sup}\ \nabla h_j(x^t) - \nabla h_{M \cup C}(x^t) \rVert^2] \right] \\
    % & = \E_{j\in J_t} \left[  \E [\underset{x}{\sup}\ \lVert \nabla h_j(x^t) - \nabla h_{M \cup C}(x^t) \rVert^2] \right] \
    & = \underset{C \subset D \subset M\cup C}\sup {\omega}^2_{h|_{D}}.
\end{align}

% \begin{align}
% \label{eq:samplevariance}
%     \underset{x}{\sup} \E \lVert \nabla g_{J_t}(x^t) - \nabla g(x^t) \rVert^2 &= \underset{x}{\sup}{n_g - b_g \over (n_g-1) b_g}\cdot {1 \over n_g} \sum_{j=1}^{n_g} \lVert \nabla g_j(x^t) - \nabla g(x^t) \rVert^2 \nonumber \\
%     &= {n_g - b_g \over (n_g-1) b_g} \sigma_g^2,
% \end{align}
% The detailed derivation is shown in technical lemma A.1 in \citep{lei2017non}.
% where $n_g$ and $b_g$ denotes the size of $C$ and minibatch $J_t$, respectively.
By the triangular inequality, we get

\begin{align}
     \E \lVert \nabla g_{J_t}(x^t) \rVert^2 &\leq  \E \lVert \nabla g_{J_t}(x^t) - \nabla h|_{M\cup C}(x^t) \rVert^2 + \E \lVert 
     \nabla h|_{M\cup C}(x^t) \rVert^2\\
    &\leq \E \lVert \nabla h|_{M \cup C}(x^t) \rVert^2+  \underset{C \subset D \subset M\cup C}\sup {\omega}^2_{h|_{D}}.
\end{align}

\end{proof}

%%%%%%%%%%%%%%%%%%%%%%%%%%%%%%%%%%%%%%%
% \begin{lemma}
% \label{thm:exp_catastrophic}
%     Let an upper bound $\beta > \beta_{H_t} >0$.
%     % The upper bound of $\Gamma_t$ 
%     For the worst case, the expectation of summing the catastrophic forgetting term over iterations $T$ is 
%     \begin{equation*}
%         \sum_{t=0}^{T-1} \Gamma_t = O(T).
%     \end{equation*}

%     % For $\delta \leq {1\over \sqrt{T}}$, we have $O(1)$.
% \end{lemma}

For continual learning, the model $x^0$ reaches to an $\epsilon$-stationary point of $f(x)$ when we have finished to learn $P$ and start to learn $C$. 
% Now, we have $\lVert \nabla f(x) \rVert = \epsilon \ll 1$
Now, we discuss the frequency of transfer and interference during continual learning before showing Lemma \ref{thm:exp_catastrophic}.
It is well known that the frequencies between interference and transfer have similar values (the frequency of constraint violation is approximately 0.5 for AGEM) as shown in Appendix \ref{sec:derivation_algo} of \citep{DBLP:conf/iclr/ChaudhryRRE19}.
Even if memory-based continual learning has a small memory buffer which contains a subset of $P$, random sampling from the buffer allows to have similar frequencies between interference and transfer.

In this paper, we consider two cases for the upper bound of $\E [\Gamma_t]$, the moderate case and the worst case. For \textbf{the moderate case}, which covers most continual learning scenarios, we assume that the inner product term $\langle \nabla f_{I_t}(x^t), \nabla g_{J_t} (x^t) \rangle$ has the same probabilities of being positive (transfer) and negative (interference).
Then, we can approximate $\E [ \langle \nabla f_{I_t}(x^t), \nabla g_{J_t} (x^t) \rangle] \approx 0$ over all randomness.
For \textbf{the worst case}, we assume that all $\langle \nabla f_{I_t}(x^t), \nabla g_{J_t} (x^t) \rangle$ has negative values.

\begin{proof}[\textbf{Proof of Lemma \ref{thm:exp_catastrophic}}]
For the moderate case, we derive the rough upper bound of $\E [\Gamma_t]$:
\begin{align}
    \E \left[\Gamma_t \right] &= \E \left[ {\beta_{H_t}^2 L \over 2} \lVert \nabla g_{J_t}(x^t) \rVert^2 - \beta_{H_t}(1-\alpha_{H_t}L) \langle \nabla f_{I_t}(x^t), \nabla g_{J_t} (x^t) \rangle\right] \\
    &\approx \E \left[ {\beta_{H_t}^2 L \over 2} \lVert \nabla g_{J_t}(x^t) \rVert^2\right] \\
    &= O \left( \E \left[ {\beta^2 L \over 2} \lVert \nabla g_{J_t}(x^t) \rVert^2 \right] \right)
\end{align}
% where $\lVert \nabla g_{J_t}(x^t) \rVert \geq \lVert \nabla f_{I_t}(x^t) \rVert$.

% {n_g - b_g \over (n_g-1) b_g}
By plugging Lemma \ref{lemma:supsigma} into $\E[\Gamma_t]$, we obtain that
\begin{align}
    \E[\Gamma_t] &\leq O \left( \E \left[ {\beta^2 L \over 2} \lVert \nabla g_{J_t}(x^t) \rVert^2 \right] \right) \\
    &= O \left( \E \left[ {\beta^2 L \over 2} \lVert \nabla h|_{M\cup C}(x^t) \rVert^2  + {\beta^2 L \over 2} \underset{C \subset D \subset M\cup C}\sup {\omega}^2_{h|_{D}}\right]\right).
    % &= O \left( E \left[ {\beta^2 L \over 2} \lVert \nabla g(x^t) \rVert^2 \right] + {\beta^2 L(n_g-b_g) \over 2(n_g -1)b_g}\sigma_g^2 \right).
\end{align}

% The sum of catastrophic forgetting term $\sum \Gamma_t$ is corrected as $\sum E[\Gamma_t]$.
We use the technique for summing up in the proof of Theorem 1,
then the cumulative sum of catastrophic forgetting term is derived as
\begin{align}
    \sum_{t=0}^{T-1} \E[\Gamma_t] &\leq  \sum_{t=0}^{T-1} {\beta^2 L \over 2}O \left( \E \left[  \lVert h|_{M\cup C}(x^t) \rVert^2 \right] +\underset{C \subset D \subset M\cup C}\sup {\omega}^2_{h|_{D}}  \right) \\
    &\leq  {\beta^2 L \over 2} \sum_{t=0}^{T-1} O \left( {1\over \beta} \left[ h|_{M\cup C}(x^t) - h|_{M\cup C}(x^{t+1}) \right] + {L\beta \over 2} \sigma_{h|_{M\cup C}}^2 +\underset{C \subset D \subset M\cup C}\sup {\omega}^2_{h|_{D}}   \right) \\
    & \leq{\beta^2 L \over 2}  O\left({1 \over \beta}\Delta_{h|_{M\cup C}} + {TL\beta \over 2} \sigma_{h|_{M\cup C}}^2 + {T\underset{C \subset D \subset M\cup C}\sup {\omega}^2_{h|_{D}}} \right) \\
    &= O\left( \beta \Delta_{h|_{M\cup C}} +   {TL \beta^3 \over 2}\sigma_{h|_{M\cup C}}^2 +T\beta^2\underset{C \subset D \subset M\cup C}\sup {\omega}^2_{h|_{D}} \right).
\end{align}
Now, we consider the randomness of memory choice.
Let $D^*$ be as follows:
\begin{align}
    D^* =   \underset{C \subset D \subset P\cup C}{\arg\max} \beta \Delta_{h|_{D}} +   {TL \beta^3 \over 2}\sigma_{h|_{D}}^2.
\end{align}
Then, we obtain the following inequality,
\begin{align}
    \sum_{t=0}^{T-1} \E[\Gamma_t] &\leq O\left( \beta \Delta_{h|_{D^*}} +   {TL \beta^3 \over 2}\sigma_{h|_{D^*}}^2 +T\beta^2\underset{C \subset D \subset M\cup C}\sup {\omega}^2_{h|_{D}} \right)\\
    &\leq O\left( \beta \Delta_{h|_{D^*}} +   {TL \beta^3 \over 2}\sigma_{h|_{D^*}}^2 +T\beta^2\underset{C \subset D \subset P\cup C}\sup {\omega}^2_{h|_{D}} \right).
\end{align}

Rearranging the above equation, we get
\begin{align}
    \sum_{t=0}^{T-1} \E[\Gamma_t] \leq O\left( T \left( {L \beta^3 \over 2}\sigma_{h|_{D^*}}^2 +\beta^2\underset{C \subset D \subset P\cup C}\sup {\omega}^2_{h|_{D}}\right) +  \beta \Delta_{h|_{D^*}} \right).
\end{align}

% Therefore, we can write $ \sum_{t=0}^{T-1} \E[\Gamma_t]=O(T)$. We note that the rough upper bound of $\sum \E[\Gamma_t]$ increases monotonically with training step as in the previous result in the paper.
\textbf{For the moderate case}, we provide the derivations of the convergence rate for two cases of $\beta$ as follows.

When $\beta < \alpha=c/\sqrt{T}$, the upper bound always satisfies
\begin{align*}
     \sum_{t=0}^{T-1} {\E[\Gamma_t] \over \sqrt{T}} &\leq {1 \over \sqrt{T}}O\left(  {1 \over T} \left( {L \beta \over 2}\sigma_{h|_{D^*}}^2 +{1\over \sqrt{T}}\underset{C \subset D \subset P\cup C}\sup {\omega}^2_{h|_{D}}\right) + {1 \over \sqrt{T}} \Delta_{h|_{D^*}}\right) < O\left( {1 \over T^{3/2}} + {1 \over T} \right).
\end{align*}

For $\beta \geq \alpha=c/\sqrt{T}$, we cannot derive a tighter bound, so we still have
\begin{align*}
     \sum_{t=0}^{T-1} {\E[\Gamma_t] \over \sqrt{T}} &\leq {1 \over \sqrt{T}}O\left(  T \left( {L \beta^3 \over 2}\sigma_{h|_{D^*}}^2 +\beta^2\underset{C \subset D \subset P\cup C}\sup {\omega}^2_{h|_{D}}\right) +  \beta \Delta_{h|_{D^*}} \right) = O\left(\sqrt{T} + {1 \over \sqrt{T}} \right).
\end{align*}
% This result is obtained by dividing $\sum E[\Gamma_t]$ by $\sqrt{T}$ as in the proof of Thm. 1.
% \textb{On the other hand, $\E[\Gamma_t]$ can be negative when $\langle \nabla f_{I_t}(x^t), \nabla g_{J_t} (x^t) \rangle >0$.
% It implies that the cumulative sum of $\E[\Gamma_t]$ does not increase monotonically.}
% \textr{By Lemma \ref{lemma:g}}
% Therefore, for some large number $N<O(T)$, we can denote the cumulative sum of $\E[\Gamma_t]$ over the finite steps $T$ as
% \begin{equation}
% \label{aeq:o1}
%     \sum_{t=0}^{T-1} {\E[\Gamma_t] \over \sqrt{T}} \leq {N\over \sqrt{T}} = O({1\over \sqrt{T}}).
% \end{equation}

\textbf{For the worst case}, we assume that there exists a constant $c_{f,g}$ which satisfies $c_{f,g} \lVert \nabla g_{J_t}(x^t) \rVert \geq  \lVert \nabla f_{I_t}(x^t) \rVert$.
\begin{align}
    \E \left[\Gamma_t \right] &= \E \left[ {\beta_{H_t}^2 L \over 2} \lVert \nabla g_{J_t}(x^t) \rVert^2 - \beta_{H_t}(1-\alpha_{H_t}L) \langle \nabla f_{I_t}(x^t), \nabla g_{J_t} (x^t) \rangle\right] \\
    &\leq \E \left[ {\beta_{H_t}^2 L \over 2} \lVert \nabla g_{J_t}(x^t) \rVert^2 + \beta_{H_t}(1-\alpha_{H_t}L) \lVert \nabla f_{I_t}(x^t)\rVert \lVert \nabla g_{J_t} (x^t) \rVert\right] \\
    &\leq \E \left[ {\beta^2 L \over 2} \lVert \nabla g_{J_t}(x^t) \rVert^2 + \beta c_{f,g}\lVert\nabla g_{J_t} (x^t) \rVert^2 \right] \\
    &= O \left( \E \left[ \left(\beta^2 + \beta\right) \lVert \nabla g_{J_t}(x^t) \rVert^2 \right] \right).
\end{align}
% where $\lVert \nabla g_{J_t}(x^t) \rVert \geq \lVert \nabla f_{I_t}(x^t) \rVert$.

% {n_g - b_g \over (n_g-1) b_g}
By plugging Lemma \ref{lemma:supsigma} into $\E[\Gamma_t]$, we obtain that
\begin{align}
    \E[\Gamma_t] &\leq O \left( \E \left[ \left(\beta^2 + \beta\right)  \lVert \nabla g_{J_t}(x^t) \rVert^2 \right] \right) \\
    &= O \left( \left(\beta^2 + \beta\right) \E \left[  \lVert \nabla h|_{M\cup C}(x^t) \rVert^2  +  \underset{C \subset D \subset M\cup C}\sup {\omega}^2_{h|_{D}}\right]\right).
    % &= O \left( E \left[ {\beta^2 L \over 2} \lVert \nabla g(x^t) \rVert^2 \right] + {\beta^2 L(n_g-b_g) \over 2(n_g -1)b_g}\sigma_g^2 \right).
\end{align}

% The sum of catastrophic forgetting term $\sum \Gamma_t$ is corrected as $\sum E[\Gamma_t]$.
We use the technique for summing up in the proof of Theorem 1,
then the cumulative sum of catastrophic forgetting term is derived as
\begin{align}
    \sum_{t=0}^{T-1} \E[\Gamma_t] &\leq  \sum_{t=0}^{T-1} \left(\beta^2 + \beta\right) O \left( \E \left[  \lVert h|_{M\cup C}(x^t) \rVert^2 \right] +\underset{C \subset D \subset M\cup C}\sup {\omega}^2_{h|_{D}}  \right) \\
    &\leq \left(\beta^2 + \beta\right) \sum_{t=0}^{T-1} O \left( {1\over \beta} \left[ h|_{M\cup C}(x^t) - h|_{M\cup C}(x^{t+1}) \right] + {L\beta \over 2} \sigma_{h|_{M\cup C}}^2 +\underset{C \subset D \subset M\cup C}\sup {\omega}^2_{h|_{D}}   \right) \\
    & \leq\left(\beta^2 + \beta\right)  O\left({1 \over \beta}\Delta_{h|_{M\cup C}} + {TL\beta \over 2} \sigma_{h|_{M\cup C}}^2 + {T\underset{C \subset D \subset M\cup C}\sup {\omega}^2_{h|_{D}}} \right) \\
    &= O\left( (\beta+1) \Delta_{h|_{M\cup C}} +   {TL \beta^2(\beta+1) \over 2}\sigma_{h|_{M\cup C}}^2 +T\beta(\beta+1)\underset{C \subset D \subset M\cup C}\sup {\omega}^2_{h|_{D}} \right).
\end{align}

For the worst case, we provide the derivations of the convergence rate for two cases of $\beta$ as follows.

When $\beta < \alpha=c/\sqrt{T}$, the upper bound always satisfies
\begin{align*}
     \sum_{t=0}^{T-1} {\E[\Gamma_t] \over \sqrt{T}} &\leq {1 \over \sqrt{T}}O\left(   {L c + \sqrt{T} \over \sqrt{T}}\sigma_{h|_{D^*}}^2 +(\sqrt{T} + c)\underset{C \subset D \subset P\cup C}\sup {\omega}^2_{h|_{D}} + {\sqrt{T} + c \over \sqrt{T}} \Delta_{h|_{D^*}}\right) < O\left( {1 \over T} + {1 \over \sqrt{T}} + 1 \right).
\end{align*}

For $\beta \geq \alpha=c/\sqrt{T}$, we cannot derive a tighter bound, so we still have
\begin{align*}
     \sum_{t=0}^{T-1} {\E[\Gamma_t] \over \sqrt{T}} &\leq {1 \over \sqrt{T}}O\left(  T \left( {L \beta^2(\beta+1) \over 2}\sigma_{h|_{D^*}}^2 +\beta(\beta+1)\underset{C \subset D \subset P\cup C}\sup {\omega}^2_{h|_{D}}\right) +  (\beta+1) \Delta_{h|_{D^*}} \right) = O\left(\sqrt{T} + {1 \over \sqrt{T}} \right).
\end{align*}

\end{proof}

Even if we consider the worst case, we still have $O(1)$ for the cumulative forgetting $\E[\Gamma_t]$ when $\beta < \alpha$.
This implies that we have the theoretical condition for control the forgetting on $f(x)$ while evolving on $C$.
In the main text, we only discuss the moderate case to emphasize $f(x)$ can be converged by the effect of transfer during continual learning, but we have also considered the worst case can be well treated by our theoretical condition by keeping the convergence of $f(x)$ over time as follows.

\begin{proof}[\textbf{Proof of Corollary \ref{coro:smallbeta}}]

By Lemma \ref{thm:exp_catastrophic}, we have 
\begin{equation*}
    \sum_{t=0}^{T-1} {\E[\Gamma_t] \over \sqrt{T}} < O\left( {1 \over T^{3/2}} + {1 \over T} \right)
\end{equation*}
for $\beta < \alpha$ for \textbf{the moderate case}.
Then, we can apply the result into RHS of the inequality in Theorem \ref{thm:min} as follows.
\begin{align*}
        \underset{t}{\min}\ \mathbb{E}  \lVert \nabla f (x^t) \rVert^2  &\leq {A \over \sqrt{T}} \left({1\over c}\left( \Delta_f +   \sum_{t=0}^{T-1}\E\left[ \Gamma_t \right] \right) +  {Lc \over 2} \sigma_{f}^2 \right) \\
        &= {A/c \over \sqrt{T}} \left( \Delta_f +  {Lc^2 \over 2} \sigma_{f}^2 \right) + {A/c \over \sqrt{T}}  \sum_{t=0}^{T-1} \E[\Gamma_t] \\
        &= O\left( {1 \over T^{3/2}} + {1 \over T} + {1 \over T^{1/2}} \right)=O\left( {1 \over \sqrt{T}} \right).
\end{align*}

In addition, we have the convergence rate of $f(x)$ for \textbf{the worst case} as follows:
\begin{align}
    \underset{t}{\min}\ \mathbb{E}  \lVert \nabla f (x^t) \rVert^2 = O(1),
\end{align}
which implies that $f(x)$ can keep the convergence while evolving on $C$.

\end{proof}

\begin{proof}[\textbf{Proof of Corollary \ref{coro:one}}]
To formulate the IFO calls, Recall that $T(\epsilon)$
\begin{equation*}
    T(\epsilon) = \min \ \{ T: \ \min  \ \E \lVert \nabla f(x^t) \rVert^2 \leq \epsilon \}.
\end{equation*}
A single IFO call is invested in calculating each step, and we now compute IFO calls to reach an $\epsilon$-accurate solution.
\begin{equation*}
    {A \over \sqrt{T}} \left({1\over c}\left( \Delta_f +   \sum_{t=0}^{T-1}\E\left[ \Gamma_t \right] \right) +  {Lc \over 2} \sigma_{f}^2 \right) \to \epsilon.
\end{equation*}
% As seen in Theorem \ref{thm:min}, NCCL has a convergence rate of
% \begin{equation}
%     O\left({\sum^{T-1}_{t=0} \Gamma_t \over \sqrt{T}} \right).
% \end{equation}
% We note that the convergence rate for the worst case is
% \begin{equation}
%     O\left(\sqrt{T} \right),
% \end{equation}
% where the given model diverges on the convergence of $f(x)$.
% Then, IFO calls are denoted as $\infty$.
When $\beta < \alpha$, we get
\begin{equation*}
    \text{IFO calls} = O\left({1\over \epsilon^2}\right).
\end{equation*}
Otherwise, when $\beta \geq \alpha$, we cannot guarantee the upper bound of stationary decreases over time. Then, we cannot compute IFO calls for this case.

% For the case of Equation \ref{aeq:o1}, we obtain the convergence rate $O(1/\sqrt{T}).$
% Thus we get $O(1/\epsilon^2)$ in this case.

% $\E \lVert \nabla f(x^t) \rVert^2=O({\sum C_t \over \sqrt{T}})$ by Theorem \ref{thm:min}.
% Then by Lemma \ref{thm:sum_catastrophic}, we have
\begin{comment}
\begin{equation*}
   \underset{t}{\min} \ \E \lVert \nabla f(x^t) \rVert^2 = O\left({\beta^2\delta \sqrt{T} \over \sqrt{T}}\right)=O(\beta^2\delta).
\end{equation*}
It implies that $\underset{t}{\min} \ \E \lVert \nabla f(x^t) \rVert^2$ is not decreasing when $1 \ll \beta^2\delta \sqrt{T}$.
Then, $x^t$ cannot reach to the stationary point.

On the other hand, $f(x)$ can be converged to the stationary point when $\beta^2\delta \leq {1 \over \sqrt{T}}$ such that
\begin{equation}
    \underset{t}{\min} \ \E \lVert \nabla f(x^t) \rVert^2 = O(\beta^2\delta)=O\left({1 \over \sqrt{T}}\right).
\end{equation}
To derive a bound for $T(\epsilon)$, we note that
\begin{align*}
    O \left( { 1 \over \sqrt{T}} \right) \leq \epsilon.
\end{align*}
Then we have
\begin{align*}
    T(\epsilon) = O\left( {1 \over \epsilon^2} \right).
\end{align*}
The IFO call is defined as $\sum_{t=1}^{T( \epsilon)} b_{f,t}$. Therefore, the IFO call is $O(1/\epsilon^2)$.
\end{comment}

\end{proof}
%%%%%%%%%%%%%%%%%%%%%%%%%%%%%%%%%%%%%%%%%%%%%%%%%%%%%%%%%%%%%%%%%%%%%%%%%%%%%%%%%%%
\section{Derivation of Equations in Adaptive Methods in Continual Learning}
\label{sec:derivation_algo}

\textbf{Derivation for A-GEM} \quad
Let the surrogate $\nabla \Tilde{g}_{J_t}(x^t)$ as
\begin{align}
    \nabla \Tilde{g}_{J_t}(x^t) = \nabla g_{J_t}(x^t) - \left\langle {\nabla f_{I_t}(x^t) \over \lVert \nabla f_{I_t}(x^t) \rVert}, \nabla g_{J_t} (x^t) \right\rangle {\nabla f_{I_t}(x^t) \over \lVert \nabla f_{I_t}(x^t) \rVert},
\end{align}
where $\alpha_{H_t} = \alpha (1 - {\langle \nabla f_{I_t}(x^t), \nabla g_{J_t} (x^t) \rangle \over \lVert \nabla f_{I_t}(x^t) \rVert^2})$ and $\beta_{H_t}=\alpha$ for Equation \ref{eq:gradupdate}.

Then, we have
\begin{align}
    \E[\Gamma_t] &= \mathbb{E}\left[{\beta_{H_t}^2 L \over 2} \lVert \nabla \Tilde{g}_{J_t}(x^t) \rVert^2 - \beta_{H_t} \langle \nabla f_{I_t}(x^t), \nabla \Tilde{g}_{J_t} (x^t) \rangle \right] \nonumber \\
    &= \mathbb{E} \left[{\beta_{H_t}^2 L \over 2} \left( \lVert \nabla g_{J_t}(x^t) \rVert^2 -2{ \langle \nabla f_{I_t}(x^t), \nabla g_{J_t} (x^t) \rangle^2 \over \lVert \nabla f_{I_t}(x^t) \rVert^2} + { \langle \nabla f_{I_t}(x^t), \nabla g_{J_t} (x^t) \rangle^2 \over \lVert \nabla f_{I_t}(x^t) \rVert^2} \right) - \beta_{H_t} \langle \nabla f_{I_t}(x^t), \nabla \Tilde{g}_{J_t}(x^t)  \rangle\right] \nonumber \\
    &= \mathbb{E} \left[{\beta_{H_t}^2 L \over 2} \left( \lVert \nabla g_{J_t}(x^t) \rVert^2 -{ \langle \nabla f_{I_t}(x^t), \nabla g_{J_t} (x^t) \rangle^2 \over \lVert \nabla f_{I_t}(x^t) \rVert^2}\right) - \beta_{H_t} \left( \langle \nabla f_{I_t}(x^t),  \nabla g_{J_t}(x^t) \rangle - \langle \nabla f_{I_t}(x^t),  \nabla g_{J_t}(x^t) \rangle \right)\right] \nonumber \\
    &= \mathbb{E} \left[{\beta_{H_t}^2 L \over 2} \left( \lVert \nabla g_{J_t}(x^t) \rVert^2 -{ \langle \nabla f_{I_t}(x^t), \nabla g_{J_t} (x^t) \rangle^2 \over \lVert \nabla f_{I_t}(x^t) \rVert^2}\right) \right].
\end{align}
Now, we compare the catastrophic forgetting term between the original value with $\nabla g_{J_t} (x^t)$ and the above surrogate.
\begin{align*}
    \mathbb{E} \left[{\beta_{H_t}^2 L \over 2} \left( \lVert \nabla g_{J_t}(x^t) \rVert^2 -{ \langle \nabla f_{I_t}(x^t), \nabla g_{J_t} (x^t) \rangle^2 \over \lVert \nabla f_{I_t}(x^t) \rVert^2}\right) \right] <  \mathbb{E}\left[{\beta_{H_t}^2 L \over 2} \lVert \nabla g_{J_t}(x^t) \rVert^2 - \beta_{H_t} \langle \nabla f_{I_t}(x^t), \nabla g_{J_t} (x^t) \rangle \right].
\end{align*}
Then, we can conclude that $\E[\Gamma_t]$ with the surrogate of A-GEM  is smaller than the original $\E[\Gamma_t]$.

\textbf{Derivation of optimal $\Gamma_t^*$ and $\beta_{H_t}^*$} \quad
For a fixed learning rate $\alpha$, we have
\begin{align*}
    0={\partial \E [\Gamma_t] \over \partial \beta_{H_t}} &= \E \left[ {\partial \Gamma_t \over \partial \beta_{H_t}} \right] \\
    &=  \E \left[ \beta_{H_t} L \lVert \nabla g_{J_t} (x^t) \rVert - (1- \alpha L) \langle \nabla f_{I_t}(x^t), \nabla g_{J_t} (x^t) \rangle \right].
\end{align*}
Thus, we obtain
\begin{align*}
    \beta_{H_t}^* = {(1-\alpha_{H_t} L)\langle \nabla f_{I_t}(x^t), \nabla g_{J_t} (x^t) \rangle \over L \lVert \nabla g_{J_t}(x^t) \rVert^2}={(1-\alpha_{H_t} L)\Lambda_{H_t} \over L \lVert \nabla g_{J_t}(x^t) \rVert^2}, \\
    \Gamma_t^* = - {(1-\alpha_{H_t} L)\langle \nabla f_{I_t}(x^t), \nabla g_{J_t} (x^t) \rangle \over 2L \lVert \nabla g_{J_t}(x^t) \rVert^2}= -{(1-\alpha_{H_t} L)\Lambda_{H_t} \over 2L \lVert \nabla g_{J_t}(x^t) \rVert^2}.
\end{align*}

% \begin{comment}
\section{Overfitting to replay Memory}
\label{sec:overfitting}
\begin{comment}
In the main text, we discussed a theoretical convergence analysis of continual learning for a smooth nonconvex finite-sum optimization problems.
The practical continual learning tasks have the restriction on full access to the entire data points of previously learned tasks.
Unlike taking expectation over $I_t \sim M$ and $M \sim P \cup C$, we have to compute on the given memory in the practical scenario.
Then, we note that $\E[B_t | M] \neq 0$.

Now we rewirte Equation \ref{aeq:thm1_result} for the worst case as

\begin{align}
    % \sup \lVert \nabla f(x) \rVert^2 &\leq \sum B_t \\
     T \sup \lVert \nabla f(x) \rVert^2  &\leq {1 \over \alpha(1-\alpha L /2 )} \left( \Delta_f + \sum \left(B_t + C_t \right) + {L \over 2} \alpha^2 \sigma_f^2 \right) \\
      \sup \lVert \nabla f(x) \rVert^2  &\leq {A \over \sqrt{T}} \left( {1 \over c} \left(\Delta_f + \sum \left(B_t + C_t \right) \right) + {Lc \over 2} \sigma_f^2 \right).
\end{align}

We note that $\sum B_t$ is a random variable, which is unpredictible, and
choosing $\nabla f_M(t) = \nabla f(x^t)$ over entire period is impossible.
Then, the cumulative sum of $B_t$ is increasing over $T$.
Therefore, we conclude that for the overfitting to memory degrades the convergence rate of NCCL empirically.
\end{comment}

In Lemma \ref{lemma:step}, we show the expectation of stepwise change of upper bound.
Now, we discuss the distribution of the upper bound by analyzing the random variable $B_t$.
As $B_t$ is computed by getting
\begin{equation*}
    B_t = (L\alpha_{H_t}^2 - \alpha_{H_t}) \langle \nabla f(x^t), e_t \rangle + \beta_{H_t} \langle \nabla g_{J_t}(x^t),e_t \rangle.
\end{equation*}
The purpose of our convergence analysis is to compute the upper bound of Equation \ref{eq:thm1},
then we compute the upper bound of $B_t$.
\begin{align*}
    B_t &\leq (L\alpha_{H_t}^2 - \alpha_{H_t}) \lVert \nabla f(x^t) \rVert \lVert e_t\rVert + \beta_{H_t} \lVert \nabla g_{J_t}(x^t)\rVert \lVert e_t \rVert.
\end{align*}
It is noted that the upper bound is related to the distribution of the norm of $e_t$.
We have already know that $\E [e_t]=0$, so we consider its variance, Var$(\lVert e_t \rVert)$ in this section.
Let us denote the number of data points of $P$ in a memory $M_0$ as $m_{P}$.
We assume that $M_0$ is uniformly sampled from $P$. 
Then the sample variance, Var$(\lVert e_t \rVert)$ is computed as
\begin{align*}
    \text{Var}(\lVert e_t \rVert) = {n_f - m_{P} \over (n_f-1) m_{P}} \sigma_f^2
\end{align*}
by the similar derivation with Equation \ref{eq:samplevariance}.
The above result directly can be applied to the variance of $B_t$.
This implies $m_t$ is a key feature which has an effect on the convergence rate.
It is noted that the larger $m_P$ has the smaller variance by applying schemes, such as larger memory.
In addition, the distributions of $e_t$ and $\nabla f_{I_t}(x^t)$ are different with various memory schemes.
Therefore, we can observe that memory schemes differ the performance even if we apply same step sizes.

% % Thm 1
% % TODO: check the coefficient of B_t
% \begin{lemma}
% %Suppose $f$ has $\sigma_f$ bounded gradient. $L \alpha_{H_t}^2 - \alpha_{H_t}^2 \leq \gamma$ for some $\gamma >0$ and 
% Suppose that Assumption \ref{assumption:lsmooth} holds and $0 < \alpha_{H_t} \leq {2 \over L}$.
% For $x^t$ updated by Algorithm \ref{alg:gni}, we have
% \begin{align}
%     &\mathbb{E}_t  \lVert \nabla f (x^t) \rVert^2 \leq  \mathbb{E}_t \left[ {1 \over \alpha_{H_t}(1-{L\over2}\alpha_{H_t})} \left(f(x^t) - f(x^{t+1}) + B_t + \Gamma_t \right) + {\alpha_{H_t} L \over 2 (1-{L\over2}\alpha_{H_t})} \sigma_{f}^2 \right].
% \end{align}
% \end{lemma}

\end{document}

%% file: math_comands.tex
\usepackage{amsmath,amsfonts,bm}

\def\eqref#1{equation~\ref{#1}}
\def\1{\bm{1}}

\DeclareMathAlphabet{\mathsfit}{\encodingdefault}{\sfdefault}{m}{sl}
\SetMathAlphabet{\mathsfit}{bold}{\encodingdefault}{\sfdefault}{bx}{n}

\def\gC{{\mathcal{C}}}

\def\gP{{\mathcal{P}}}

\newcommand{\E}{\mathbb{E}}